# *Defeasible Logic Programming*
# *An Argumentative Approach*

Alejandro J. García     Guillermo R. Simari

*agarcia@cs.uns.edu.ar*     *grs@cs.uns.edu.ar*
*Departament of Computer Science & Engineering*
*Universidad Nacional del Sur - Bahía Blanca - Argentina*

**Abstract**

The work reported here introduces Defeasible Logic Programming (DeLP), a formalism that combines results of Logic Programming and Defeasible Argumentation. DeLP provides the possibility of representing information in the form of *weak rules* in a declarative manner, and a defeasible argumentation inference mechanism for warranting the entailed conclusions.

In DeLP an argumentation formalism will be used for deciding between contradictory goals. Queries will be supported by arguments that could be defeated by other arguments. A query $q$ will succeed when there is an argument $\mathcal{A}$ for $q$ that is warranted, i.e. the argument $\mathcal{A}$ that supports $q$ is found undefeated by a warrant procedure that implements a dialectical analysis.

The defeasible argumentation basis of DeLP allows to build applications that deal with incomplete and contradictory information in dynamic domains. Thus, the resulting approach is suitable for representing agent's knowledge and for providing an argumentation based reasoning mechanism to agents.

## 1 Introduction

Research in Nonmonotonic Reasoning, Logic Programming, and Argumentation have provided important results seeking to develop more powerful tools for knowledge representation and common sense reasoning. Advances in those areas are leading to important and useful results for other areas such as the development of intelligent agents and multi-agent system applications.

The work reported here introduces Defeasible Logic Programming (DeLP), a formalism that combines results of Logic Programming and Defeasible Argumentation. DeLP provides the possibility of representing information in the form of *weak rules* in a declarative manner, and a defeasible argumentation inference mechanism for warranting the entailed conclusions. These weak rules are the key element for introducing *defeasibility* (Pollock, 1995) and they will be used to represent a relation between pieces of knowledge that could be *defeated* after all things are considered. We believe that common sense reasoning should be defeasible in a way that is not explicitly programmed. Defeat should be the result of a global consideration of the corpus of knowledge of the agent performing the inference. Defeasible Argumentation provides the tools for doing this.

Defeasible Argumentation is a relatively young area, but already mature enough to provide solutions for other areas. Argumentative Systems are now being applied for developing applications in legal systems, negotiation among agents, decision making, etc, (Prakken & Vreeswijk, 2000; Chesñevar *et al.*, 2000; García *et al.*, 2000).

As we will show below, DeLP considers two kinds of program rules: *defeasible rules* used for representing weak or tentative information, like "a mammal does not fly", denoted $\sim\!\mathit{flies} \prec \mathit{mammal}$, and *strict rules* used for representing strict (sound) knowledge, like $\mathit{mammal} \leftarrow \mathit{dog}$, "a dog is a mammal". Syntactically, the symbol "$\prec$" is all that distinguishes a defeasible rule from a strict one. Pragmatically, a defeasible rule is used to represent defeasible knowledge, i.e., tentative information that may be used if nothing could be posed against it.

Defeasible rules will add a new representational capability for expressing a weaker link between the head and the body in a rule. A defeasible rule "$\mathit{Head} \prec \mathit{Body}$" is understood as expressing that "*reasons to believe in the antecedent Body provide reasons to believe in the consequent Head*" (Simari & Loui, 1992). *Strong negation* is allowed in the head of program rules, and hence may be used to represent contradictory knowledge.

Although strong negation has been introduced in several extensions of Logic Programming, many of those extensions handle contradictory programs in a way we feel could be improved. DeLP incorporates an argumentation formalism for the treatment of contradictory knowledge. This formalism allows the identification of the pieces of knowledge that are in contradiction. A dialectical process is used for deciding which information prevails. In particular, the argumentation based definition of the inference relation makes it possible to incorporate a treatment of preferences in an elegant way. In DeLP a query $q$ will succeed when there is an *argument* $\mathcal{A}$ for $q$ that is *warranted*. Intuitively, an argument is a minimal set of rules used to derive a conclusion. The warrant procedure involves looking for *counter-arguments* that could defeat $\mathcal{A}$.

Donald Nute in (Nute, 1994) remarks *"An inference is defeasible if it can be blocked or defeated in some way"*. Weak rules provide the locus where the blocking or defeating might occur. A query $q$ will succeed if a supporting argument $\mathcal{A}$ for $q$ is not defeated. In order to establish whether $\mathcal{A}$ is a non-defeated argument, *argument rebuttals* or *counter-arguments* that could be *defeaters* for $\mathcal{A}$ are considered, i.e., counter-arguments that by some criterion, are preferred to $\mathcal{A}$. Our defeaters will take the form of arguments, therefore defeaters for the defeaters may exist. In this manner, starting with an argument $\mathcal{A}$ for a query $q$, a *dialectical analysis* it will consider all the defeaters for $\mathcal{A}$, and then the defeaters for each defeater, and so on.

DeLP provides a *warrant procedure* that implements that dialectical analysis. Thus, a query $q$ will be *warranted*, if the argument $\mathcal{A}$ that supports $q$ is found undefeated by the warrant procedure. During the dialectical analysis certain constraints are imposed for averting problematic situations such as producing an infinite sequence of defeaters. Thus, DeLP can manage defeasible reasoning and handle contradictory programs, allowing the representation of defeasible and non-defeasible knowledge. We will also discuss briefly in this paper how to extend DeLP for considering default negation.

The resulting approach is suitable for representing agent's knowledge and for providing an argumentation based reasoning mechanism to agents. In (García *et al.*, 2000), a particular application for a multi-agent system in the stock market domain has been developed using DeLP. There, agents knowledge is specified in the form of a defeasible logic program, and agents reason using DeLP in order to reach decisions about buying or selling stocks.

As Brewka states in (Brewka, 2001a), argumentation plays a central role in the communication of human and artificial agents and is an ubiquitous task in professional and everyday life. We agree with Brewka that it is necessary to "...*take not only the logical, but also the procedural character of argumentation seriously...*". It is remarkable that Nute's Defeasible Logic (Nute, 1994), recent extensions to defeasible logic (Antoniou *et al.*, 2000a; Antoniou *et al.*, 1998), and several defeasible argumentation formalisms (Loui, 1997a; Prakken & Sartor, 1997; Vreeswijk, 1997), also consider defeasible rules for representing knowledge. However, in most of these formalisms, a priority relation among rules must be explicity given with the program in order to decide between rules with contradictory consequents.

This work is organized as follows. Section 2 introduces the language of DeLP without default negation. In Section 3 the defeasible argumentation formalism is developed. and two comparison criteria are introduced. A dialectical procedure for obtaining a warrant conclusion is developed in Section 5. In Section 6 DeLP with default negation is considered. Implementation issues and applications are described in Section 7, and finally in Section 8 the related work is surveyed.

## 2 The Language

The DeLP language is defined in terms of three disjoint sets: a set of *facts*, a set of *strict rules* and a set of *defeasible rules*. In the language of DeLP a literal "$L$" is a ground atom "$A$" or a negated ground atom "$\sim A$", where "$\sim$" represents the *strong negation*. Hence, literals have no variables.

**Definition** *2.1* (*Fact*)
A fact is a literal, i.e. a ground atom, or a negated ground atom.

**Definition** *2.2* (*Strict Rule*)
A Strict Rule is an ordered pair, denoted "*Head* $\leftarrow$ *Body*", whose first member, *Head*, is a literal, and whose second member, *Body*, is a finite non-empty set of literals. A strict rule with the head $L_0$ and body $\{L_1, \ldots, L_n\}$ can also be written as: $L_0 \leftarrow L_1, \ldots, L_n$ ($n > 0$).

**Definition** *2.3* (*Defeasible Rule*)
A Defeasible Rule is an ordered pair, denoted "*Head* $\prec$ *Body*", whose first member, *Head*, is a literal, and whose second member, *Body*, is a finite non-empty set of literals. A defeasible rule with head $L_0$ and body $\{L_1, \ldots, L_n\}$ can also be written as: $L_0 \prec L_1, \ldots, L_n$ ($n > 0$.)

The syntax of strict rules correspond to *basic rules* (Lifschitz, 1996), but we call them 'strict' to emphasize the difference with the 'defeasible' ones. Observe that strong negation may be used in the head of the rules. Some examples of strict rules are: "$\sim$*innocent* $\leftarrow$ *guilty*", "*dead* $\leftarrow$ $\sim$*alive*".

Syntactically, the symbol "—<" is all that distinguishes a defeasible rule from a strict one. Pragmatically, a defeasible rule is used to represent defeasible knowledge, i.e., tentative information that may be used if nothing could be posed against it. Thus, whereas a strict rule is used to represent non-defeasible information such as: "*bird* $\leftarrow$ *penguin*" which expresses that *"all penguins are birds"*, a defeasible rule is used to represent defeasible knowledge such as "*flies* —< *bird*" which expresses that: *"reasons to believe that it is a bird provide reasons to believe that it flies"*, or *"birds are presumed to fly"* or *"usually, a bird can fly."*

The symbols " —< " and " $\leftarrow$ " denote meta-relations between sets of literals, and have no interaction with language symbols. In particular, there is no contraposition for program rules. In our examples, we will follow standard typographic conventions of Logic Programming extending them conveniently for representing DeLP rules.

Program rules have a non-empty body. Observe that a strict rule with an empty body can be represented as a fact, but a defeasible rule with an empty body is not a fact. For example: "$a$ —< " would express that *"there are (defeasible) reasons to believe in $a$"*, and since this information is defeasible, then $a$ could not be a fact. A defeasible rule with an empty body was introduced in several approaches and is called a *presumption* (Nute, 1988; García *et al.*, 1998; García, 2000). For technical reasons we will develop the core of DeLP without considering presumptions, and they will be added as an extension to DeLP (see Section 6).

Defeasible rules are not default rules. In a default rule $\varphi : \psi_1 \ldots, \psi_n/\chi$ the *justification* part $\psi_1 \ldots, \psi_n$ is a consistency check that contributes in the control of the applicability of the rule. A defeasible rule represents a weak connection between head and body of the rule. The effect of a defeasible rule comes from a dialectical analysis, made by the inference mechanism, which involves the consideration of arguments and counter-arguments where that rule is included. Therefore, in a defeasible rule there is no need to encode any particular check. The advantage of this is notorious when the beliefs supported by knowledge represented by default rules change. The changes will lead to re-representing the new beliefs by modifying the justification part of those rules, which could lead to a re-representation cascade effect, see (Brewka & Eiter, 2000). Changes in the knowledge represented by defeasible rules, in general, is reflected with the sole addition of new defeasible rules to the representation. Therefore, if the knowledge base changes frequently, defeasible rules are a better alternative.

In Example 8.1, we will introduce a more illustrative example comparing defaults and defeasible rules. The interested reader is referred to (Nute, 1994) where a comparison with other nonmonotonic theories is given.

**Definition** *2.4* (*Defeasible Logic Program*)
A Defeasible Logic Program $\mathcal{P}$, abbreviated *de.l.p.*, is a possibly infinite set of facts, strict rules and defeasible rules. In a program $\mathcal{P}$, we will distinguish the subset $\Pi$ of facts and strict rules, and the subset $\Delta$ of defeasible rules. When required, we will denote $\mathcal{P}$ as $(\Pi, \Delta)$.

Strict and defeasible rules are ground. However, following the usual convention (Lifschitz, 1996), some examples will use "schematic rules" with variables. Given a "schematic rule" $R$, $Ground(R)$ stands for the set of all ground instances of $R$. Given a *de.l.p.* $\mathcal{P}$ with schematic rules, we define (Lifschitz, 1996):

$$Ground(\mathcal{P}) = \bigcup_{R \in \mathcal{P}} Ground(R)$$

In order to distinguish variables from other elements of a schematic rule, we will denote variables with an initial uppercase letter.

In DeLP there are four possible answers for a query: YES, NO, UNDECIDED, or UNKNOWN (see Definition 5.3). Next, we will introduce some examples of defeasible logic programs, anticipating what will be the results of using our approach. In the following sections we will explain how the mentioned results are obtained.

**Example** *2.1*
Here follows the *de.l.p.* $\mathcal{P}_{2.1} = (\Pi, \Delta)$, where sets $\Pi$ and $\Delta$ have been separated for convenience of the presentation:

$$\Pi = \left\{ \begin{array}{l} bird(X) \leftarrow chicken(X) \\ bird(X) \leftarrow penguin(X) \\ \sim flies(X) \leftarrow penguin(X) \\ chicken(tina) \\ penguin(tweety) \\ scared(tina) \end{array} \right\} \quad \Delta = \left\{ \begin{array}{l} flies(X) \prec bird(X) \\ \sim flies(X) \prec chicken(X) \\ flies(X) \prec chicken(X), scared(X) \\ nests\_in\_trees(X) \prec flies(X) \end{array} \right\}$$

As will show in the following sections, in DeLP the answer for *flies(tina)* will be YES, whereas the answer for $\sim$*flies(tina)* will be NO. The answer for *flies(tweety)* will be NO, whereas the answer for $\sim$*flies(tweety)* will be YES. ∎

**Example** *2.2*
Consider the following *de.l.p.* adapted from (Prakken & Vreeswijk, 2000):

$$\mathcal{P}_{2.2} = \left\{ \begin{array}{l} has\_a\_gun(X) \prec lives\_in\_chicago(X) \\ \sim has\_a\_gun(X) \prec lives\_in\_chicago(X), pacifist(X) \\ lives\_in\_chicago(nixon) \\ pacifist(X) \prec quaker(X) \\ \sim pacifist(X) \prec republican(X) \\ quaker(nixon) \\ republican(nixon) \end{array} \right\}$$

As we will show in Example 5.4, the expected result will be obtained: the answer for *pacifist(nixon)* will be UNDECIDED, and the answer for $\sim$*pacifist(nixon)* will be also UNDECIDED. In a case like this, other approaches have the problem of the propagation of that "indecision", and the answer for *has_a_gun(nixon)* is also undecided. However, this result will not happen in DeLP (see Example 5.5). ∎

**Example** *2.3*
Consider the following *de.l.p.* $\mathcal{P}_{2.3} = (\Pi, \Delta)$:

$$\Pi = \left\{ \begin{array}{l} h \leftarrow a \\ \sim h \leftarrow c \\ b \\ d \end{array} \right\}$$

$$\Delta = \left\{ \begin{array}{l} a \prec b \\ c \prec d \end{array} \right\}$$

In several approaches the literal $a$ is accepted as proved from $\mathcal{P}_{2.3}$ because there is no rule with "$\sim a$" in its head, so no rule that contradicts "$a \prec b$" can be found. However, observe that from literals "$a$" and "$c$" and the strict rules "$h \leftarrow a$" and "$\sim h \leftarrow c$" two contradictory literals can be derived. In DeLP, this kind of contradiction through strict rules will be discovered and the answer for the literals "$a$" and "$c$" will be UNDECIDED. See section 8 for further details. ∎

**Example** *2.4*
The following de.l.p. represents some information in the stock market domain:

$$\mathcal{P}_{2.4} = \left\{ \begin{array}{l} buy\_stock(T) \prec good\_price(T) \\ \sim buy\_stock(T) \prec good\_price(T), risky\_company(T) \\ risky\_company(T) \prec in\_fusion(T, Y) \\ risky\_company(T) \prec closing(T) \\ \sim risky\_company(T) \prec in\_fusion(T, Y), strong(Y) \\ good\_price(acme) \\ in\_fusion(acme, steel) \\ strong(steel) \end{array} \right\}$$

As we will explain next, here the answer for $buy\_stock(acme)$ will be YES. ∎

We will define next what constitute a defeasible derivation. In the following sections a proof procedure based on a defeasible argumentation formalism will be defined.

**Definition** *2.5* (*Defeasible Derivation*)
Let $\mathcal{P} = (\Pi, \Delta)$ be a *de.l.p.* and $L$ a ground literal. A defeasible derivation of $L$ from $\mathcal{P}$, denoted $\mathcal{P} \mathrel{|\!\sim} L$, consists of a finite sequence $L_1, L_2, \ldots, L_n = L$ of ground literals, and each literal $L_i$ is in the sequence because:

(a) $L_i$ is a fact in $\Pi$, or
(b) there exists a rule $R_i$ in $\mathcal{P}$ (strict or defeasible) with head $L_i$ and body $B_1, B_2, \ldots, B_k$ and every literal of the body is an element $L_j$ of the sequence appearing before $L_i$ ($j < i$.)

Given a *de.l.p.* $\mathcal{P}$, a derivation for a literal $L$ from $\mathcal{P}$ is called 'defeasible', because as we will show next, there may exist information in contradiction with $L$ that will prevent the acceptance of $L$ as a valid conclusion. If the program $\mathcal{P}$ is expressed using schematic rules, then $Ground(\mathcal{P})$ is used for obtaining defeasible derivations.

**Example** *2.5*

*Considering Example 2.1, the sequence:*

$$chicken(tina), bird(tina), flies(tina),$$

*is a defeasible derivation for the literal "flies(tina)", obtained from the following set of rules in $\mathcal{P}_{2.1}$: { (bird(tina) $\leftarrow$ chicken(tina)), (flies(tina) $\prec$ bird(tina)) }*[1]*.*

*Observe that from $\mathcal{P}_{2.1}$, there exists also a defeasible derivation for $\sim$flies(tina) from the sequence: chicken(tina), $\sim$flies(tina).* ∎

**Observation** *2.1*
*Defeasible derivation is monotonic, i.e., let $\mathcal{P}$ be a de.l.p. and $R$ be a set of program rules, if $h$ has a defeasible derivation from $\mathcal{P}$, then $h$ has a defeasible derivation from $\mathcal{P} \cup R$.*

**Observation** *2.2*
*If a program $\mathcal{P}$ has no facts, then no defeasible derivation can be obtained.*

Given a literal $L$, there could be more than one defeasible derivation for $L$. Observe also, that a defeasible derivation could use both defeasible and strict rules, or it could use only one kind of rule. In what follows, sometimes we will call *strict derivation*, a derivation where only strict rules, and facts, are used. For instance, the literal "$\sim$flies(tweety)" has a strict derivation from $\mathcal{P}_{2.1}$. The formal definition follows.

**Definition** *2.6* (*Strict Derivation*)
Let $\mathcal{P}$ be a *de.l.p.* and $h$ a literal with a defeasible derivation $L_1, L_2, \ldots, L_n = h$. We will say that $h$ has a *strict derivation* from $\mathcal{P}$, denoted $\mathcal{P} \vdash L$, if either $h$ is a fact or all the rules used for obtaining the sequence $L_1, L_2, \ldots, L_n$ are strict rules.

The symbol "$\overline{\phantom{p}}$" will be used to denote the complement of a literal with respect to strong negation, i.e. $\overline{p}$ is $\sim p$, and $\overline{\sim p}$ is $p$. Two literals are contradictory if they are complementary. Thus, *flies(tina)* and $\sim$*flies(tina)* are contradictory literals.

**Definition** *2.7* (*Contradictory set of rules*)
A set of rules is *contradictory* if and only if, there exists a defeasible derivation for a pair of complementary literals from this set.

Observe that the *de.l.p.* of Example 2.1 is a contradictory set of rules because both *flies(tina)* and $\sim$*flies(tina)* can be defeasibly derived, (see Example 2.5.) The same happens to the *de.l.p.* of Example 2.4 for *buy_stock(acme)* and $\sim$*buy_stock(acme)*.

The use of strong negation in program rules enriches language expressiveness, and also allows to represent contradictory knowledge. In general, a useful defeasible logic program will be a contradictory set of rules. However, the set $\Pi$ of facts and strict rules in a defeasible logic program $\mathcal{P}$, which is used to represent non-defeasible information, must posses certain internal coherence.

**Observation** *2.3*

---

[1] When required, parenthesis will be used for distinguishing one rule from another.

*Note that the set $\Pi$ of a de.l.p. corresponds to a logic program with strong negation. If a contradictory set $\Pi$ is used in a de.l.p. then the answer would be Lit, as in Extended Logic Programming. Therefore, we will have the convention that in a de.l.p. $\mathcal{P}$ the set $\Pi$ is non-contradictory.*

Although the set $\Pi$ is assumed to be non-contradictory, $\mathcal{P}$ itself (i.e., $\Pi \cup \Delta$), could be contradictory. It is only in this form that a *de.l.p.* may contain contradictory information. Observe that the set $\Pi$ of $\mathcal{P}_{2.1}$ is non-contradictory, whereas the whole program $\mathcal{P}_{2.1}$ is contradictory: the literals "*flies(tina)*" and "$\sim$*flies(tina)*" have both a defeasible derivation from $\mathcal{P}_{2.1}$.

In a case like this, when contradictory goals could be defeasibly derived, a formalism for deciding between them is needed. DeLP uses a defeasible argumentation formalism in order to perform such a task.

Strong negation was introduced in Extended Logic Programming (Gelfond & Lifschitz, 1990). In this formalism, when a pair of contradictory literals is derived, the set *Lit* of all literals is derived without considering any further analysis. Suppose that we regard Example 2.1 above as an Extended Logic Program, by considering all its rules as strict rules. This program would be an Extended Logic Program without "*not*", and clearly, from that program a pair of complementary literals (*"flies(tina)"* and *"$\sim$flies(tina)"*) could be derived. Therefore, the answer set calculated according to (Gelfond & Lifschitz, 1990) is *Lit*.

For deciding between contradictory goals, other formalism use a priority relation among program rules that need to be included in the program (Nute, 1994; Antoniou *et al.*, 2000a; Antoniou *et al.*, 1998; Prakken & Sartor, 1997; Kakas *et al.*, 1994; Dimopoulos & Kakas, 1995) (see Section 8 below.)

In DeLP, no priority relation is needed for deciding between contradictory goals. This characteristic maintains the declarative nature of the knowledge represented in DeLP, i.e. the interaction among the *pieces* of knowledge is not expressed *in* the language in any way but as a result of the influence of the whole corpus of the agent's knowledge. For that reason, the burden of the defeasible inference falls upon the *language processor*, i.e. our system, which figures out the interactions, instead of on the *knowledge encoder*, i.e., the programmer. The programmer does not have to evoke the *behavior* of the representation in order to add procedural control to the defeasible rules. However, as it will be shown in Section 3.2.2, priorities between defeasible rules could be used in DeLP as an alternative comparison approach.

## 3 Defeasible Argumentation

In this section, a defeasible argumentation formalism will be introduced. This formalism evolved from (Simari & Loui, 1992; Simari *et al.*, 1994b; García *et al.*, 1998) into the DeLP framework. The central notion of the formalism is the notion of *argument*. Informally, an argument is a minimal and non-contradictory set of rules used to derive a conclusion. In DeLP, answers to queries will be supported by an argument. Thus, although a *de.l.p.* could be contradictory, answers to queries will be supported by a non-contradictory set of rules. The formal definition follows.

**Definition** *3.1* (*Argument Structure*)
Let $h$ be a literal, and $\mathcal{P}=(\Pi, \Delta)$ a *de.l.p.*. We say that $\langle \mathcal{A}, h \rangle$ is an argument structure for $h$, if $\mathcal{A}$ is a set of defeasible rules of $\Delta$, such that:

1. there exists a defeasible derivation for $h$ from $\Pi \cup \mathcal{A}$,
2. the set $\Pi \cup \mathcal{A}$ is non-contradictory, and
3. $\mathcal{A}$ is minimal: there is no proper subset $\mathcal{A}'$ of $\mathcal{A}$ such that $\mathcal{A}'$ satisfies conditions (1) and (2).

In short, an argument structure $\langle \mathcal{A}, h \rangle$, or simply an *argument* $\mathcal{A}$ for $h$, is a minimal non-contradictory set of defeasible rules, obtained from a defeasible derivation for a given literal $h$. The literal $h$ will also be called the 'conclusion' supported by $\mathcal{A}$. This notion was adapted from the *argument structure* definition given in (Simari & Loui, 1992). Note that strict rules are not part of an argument structure.

**Example** *3.1*
*Consider the defeasible logic program of Example 2.1. The literal "$\sim$flies(tina)" is supported by the following argument structure:*

$$\langle \{\sim\!\mathit{flies}(\mathit{tina}) \prec \mathit{chicken}(\mathit{tina})\}, \sim\!\mathit{flies}(\mathit{tina}) \rangle$$

*whereas the literal "flies(tina)" has two argument structures that support it:*

$$\langle \{\mathit{flies}(\mathit{tina}) \prec \mathit{bird}(\mathit{tina})\}, \mathit{flies}(\mathit{tina}) \rangle$$
$$\langle \{\mathit{flies}(\mathit{tina}) \prec \mathit{chicken}(\mathit{tina}), \mathit{scared}(\mathit{tina})\}, \mathit{flies}(\mathit{tina}) \rangle$$

■

Observe that in DeLP the construction of argument structures is nonmonotonic. That is, adding facts or strict rules to the program may cause some argument structures to be invalidated because they become contradictory. For example, consider the program $\mathcal{P}=(\Pi, \Delta)$, where $\Pi=\{a\}$, and $\Delta=\{h \prec a\}$. From $\mathcal{P}$ there is an argument structure $\langle \mathcal{A}, h \rangle = \langle \{h \prec a\}, h \rangle$. However, from $\Pi \cup \{\sim h\}$, $\langle \mathcal{A}, h \rangle$ is not longer an argument structure because the set $\Pi \cup \{\sim h\} \cup \{h \prec a\}$ is a contradictory set. In other approaches to defeasible argumentation, argument construction is monotonic, and contradictory arguments are considered 'self-defeating' arguments, that is, arguments that defeat themselves. Condition 2 of the previous definition prevents the occurrence of self-defeating argument structures. We will come back to this issue in Section 4.1.

**Observation** *3.1*
*If there exists a strict derivation for $q$, then there exists a unique argument structure for $q$: $\langle \emptyset, q \rangle$. This is so because there exists a defeasible derivation for $q$ from $\Pi \cup \emptyset$, $\Pi \cup \emptyset$ is not contradictory by Observation 2.3, and there is no subset of $\emptyset$. The argument structure $\langle \emptyset, q \rangle$ is unique because by condition 3 of Definition 3.1 (minimality) no superset of $\emptyset$ can be an argument structure. For example, the literal "$\sim$flies(tweety)" has a strict derivation from $\mathcal{P}_{2.1}$, so $\langle \emptyset, \sim\!\mathit{flies}(\mathit{tweety}) \rangle$ is its the unique argument structure.*

It is interesting to note, that if $q$ has a strict derivation, then $q$ has a derivation from $\Pi$; hence, $q$ has a defeasible derivation from $\Pi \cup \mathcal{A}$, for any set $\mathcal{A}$ of defeasible rules (see Observation 2.1.) Therefore, if there exists a defeasible derivation for $\overline{q}$ from $\Pi \cup \mathcal{A}$, for some set $\mathcal{A}$, then $\Pi \cup \mathcal{A}$ will be contradictory, and no argument structure for $\overline{q}$ could be obtained, because condition 2 of the definition of argument structure will not be satisfied.

**Definition** *3.2* (*Sub-argument structure*)
An argument structure $\langle \mathcal{B}, q \rangle$ is a *sub-argument structure* of $\langle \mathcal{A}, h \rangle$ if $\mathcal{B} \subseteq \mathcal{A}$.

It is important to note that the union of arguments is not always an argument. That is, given two argument structures $\langle \mathcal{A}, h \rangle$ and $\langle \mathcal{B}, q \rangle$, the set $\mathcal{A} \cup \mathcal{B}$ could be improper for being used as an argument, because $\mathcal{A} \cup \mathcal{B}$ could be not minimal or $\mathcal{A} \cup \mathcal{B} \cup \Pi$ could be contradictory (see the following example.)

**Example** *3.2*
Consider the de.l.p.:

$$\left\{ \begin{array}{lll} b \prec c & b \prec d & h \leftarrow h_1, h_2 \\ c & d & p \leftarrow h_1 \\ h_1 \leftarrow b & h_2 \leftarrow b & \sim p \leftarrow h_2 \end{array} \right\}$$

The following argument structures can be obtained: $\langle \mathcal{A}_1, h_1 \rangle = \langle \{b \prec c\}, h_1 \rangle$ and $\langle \mathcal{A}_2, h_2 \rangle = \langle \{b \prec d\}, h_2 \rangle$. Consider now the set $\mathcal{A} = \mathcal{A}_1 \cup \mathcal{A}_2 = \{(b \prec c), (b \prec d)\}$. From $\Pi \cup \mathcal{A}$ there exists a defeasible derivation for $h$, however, $\langle \mathcal{A}, h \rangle$ is not an argument structure because $\Pi \cup \mathcal{A}$ is contradictory since $p$ and $\sim p$ have a defeasible derivation from $\Pi \cup \mathcal{A}$. Observe also that $\mathcal{A}$ is not the minimal set of defeasible rules that provides an argument structure for $h$ because $\mathcal{A}_1$ is a proper subset of $\mathcal{A}$, and $\langle \mathcal{A}, h \rangle$ is an argument structure. ■

As shown in Example 3.1, it is possible to have argument structures for contradictory literals. Thus, when considering a literal $q$, supported by argument $\mathcal{A}$, arguments that contradict $\mathcal{A}$ could exist. In that situation it is clear that it would be interesting to have a preference criterion for deciding among arguments and their counter-arguments. This issue will be explored in the next section, but first, we will introduce a graphical representation for arguments that will be used in what follows.

Arguments will be depicted as triangles. The upper vertex of the triangle will be labeled with the argument's conclusion, and the set of defeasible rules in the argument will be associated with the triangle itself. Sub-arguments will be represented as smaller triangles contained in the triangle which corresponds to the main argument at issue. Figure 1 shows the graphical representation of an argument $\langle \mathcal{A}, h \rangle$, and one of its sub-arguments $\langle \mathcal{B}, q \rangle$.

### *3.1 Rebuttals or Counter-Arguments*

In DeLP, answers to queries will be supported by arguments. However, an argument may be *defeated* by other arguments. Informally, a query $q$ will succeed if the supporting argument for it is not defeated. In order to establish whether $\mathcal{A}$ is a non-defeated argument, *argument rebuttals* or *counter-arguments* that could be *defeaters* for $\mathcal{A}$ are considered, i.e., counter-arguments that for some criterion, are preferred to $\mathcal{A}$. Since counter-arguments are arguments, there may exist defeaters for them, and so on. That suggests a dialectical analysis, that will be formally introduced in the next sections. Here, the notion of rebuttal or counter-argument is introduced.

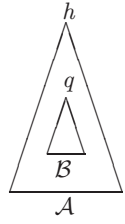

Fig. 1. An argument $\langle \mathcal{A}, h \rangle$, and a sub-argument $\langle \mathcal{B}, q \rangle$

**Definition** *3.3* (*Disagreement*)
Let $\mathcal{P} = \Pi \cup \Delta$ be a *de.l.p.*. We say that two literals $h$ and $h_1$ *disagree*, if and only if the set $\Pi \cup \{h, h_1\}$ is contradictory.

Two complementary literals $p$ and $\sim p$ trivially disagree since for any set $\Pi$, $\{p, \sim p\} \cup \Pi$ is contradictory. Furthermore, two literals that are not complementary can also disagree. For example given $\Pi = \{(\sim h \leftarrow b), (h \leftarrow a)\}$, the literals $a$ and $b$ disagree because $\Pi \cup \{a, b\}$ is contradictory. We will show below how this notion of disagreement will allow us to find direct and indirect conflicts between arguments.

**Definition** *3.4* (*Counter-argument*)
We say that $\langle \mathcal{A}_1, h_1 \rangle$ *counter-argues*, *rebuts*, or *attacks* $\langle \mathcal{A}_2, h_2 \rangle$ at literal $h$, if and only if there exists a sub-argument $\langle \mathcal{A}, h \rangle$ of $\langle \mathcal{A}_2, h_2 \rangle$ such that $h$ and $h_1$ *disagree*.

If $\langle \mathcal{A}_1, h_1 \rangle$ counter-argues $\langle \mathcal{A}_2, h_2 \rangle$ at literal $h$, then $h$ is called a *counter-argument point*, and the sub-argument $\langle \mathcal{A}, h \rangle$ is called the *disagreement sub-argument*. Figure 2-(left) gives a graphical representation of an argument and one of its counter-arguments in the conditions of Definition 3.4.

**Example** *3.3*
Continuing with Example 3.1, $\langle \{\sim\!flies(tina) \prec chicken(tina)\}, \sim\!flies(tina) \rangle$ is a counter-argument for $\langle \{flies(tina) \prec bird(tina)\}, flies(tina) \rangle$ and vice versa. Note that in this particular case, the disagreement sub-argument is the argument itself. Consider now the argument structure $\langle \mathcal{A}_4, nests\_in\_trees(tina) \rangle$, where

$$\mathcal{A}_4 = \left\{ \begin{array}{l} nests\_in\_trees(tina) \prec flies(tina) \\ flies(tina) \prec bird(tina) \end{array} \right\}$$

The argument structure $\langle \{\sim\!flies(tina) \prec chicken(tina)\}, \sim\!flies(tina) \rangle$ is a counter-argument for $\langle \mathcal{A}_4, nests\_in\_trees(tina) \rangle$ attacking an inner point, i.e., the counter-argument point is $flies(tina)$. ∎

As shown in the previous Example, a counter-argument $\langle \mathcal{A}_1, h_1 \rangle$ for $\langle \mathcal{A}_2, h_2 \rangle$, can attack directly the conclusion $h_2$ of $\langle \mathcal{A}_2, h_2 \rangle$, or can attack an inner point $h$ of $\langle \mathcal{A}_2, h_2 \rangle$. Hence, we will distinguish between a "direct" attack and an "indirect" attack. Figure 2 shows both cases.

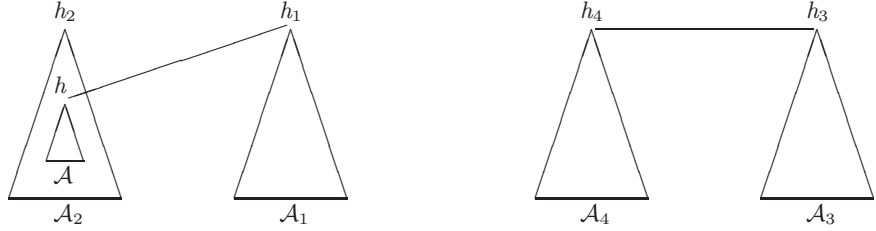

Fig. 2. Indirect attack (left) and direct attack (right)

It is interesting to remark the following property of the counter-argument notion. If $\langle \mathcal{A}_3, h_3 \rangle$ is a counter-argument for $\langle \mathcal{A}_4, h_4 \rangle$ and the counter-argument point is $h_4$ (direct attack) then $\langle \mathcal{A}_4, h_4 \rangle$ is also a counter-argument for $\langle \mathcal{A}_3, h_3 \rangle$ (see Figure 2-right.) In the case that $\langle \mathcal{A}_1, h_1 \rangle$ is a counter-argument for $\langle \mathcal{A}_2, h_2 \rangle$ attacking an inner point $h$, with $\langle \mathcal{A}, h \rangle$ as the disagreement sub-argument, then $\langle \mathcal{A}, h \rangle$ is a counter-argument for $\langle \mathcal{A}_1, h_1 \rangle$ (see Figure 2-left.)

In the example above, the argument and its counter-argument have complementary conclusions (*flies*(*tina*) and $\sim$*flies*(*tina*).) In the following example we show an argument and a counter-argument supporting literals that are not complementary. The disagreement arises through $\Pi$.

**Example** *3.4*
Consider again $\mathcal{P}_{2.3}=(\Pi, \Delta)$, $\Pi = \{ (h \leftarrow a), b, d, (\sim h \leftarrow c) \}$ and $\Delta = \{ (a \prec b), (c \prec d) \}$. From this program $\langle \{a \prec b\}, a \rangle$ is a counter-argument for $\langle \{c \prec d\}, c \rangle$, because literals $a$ and $c$ disagree.  ∎

In Observation 3.1, we have shown that any literal $q$ that has a strict derivation is supported by $\langle \emptyset, q \rangle$. The following proposition states that for any *de.l.p.* $\mathcal{P}$, and any literal $q$, an argument structure $\langle \emptyset, q \rangle$ will not have any counter-arguments. In other words, strictly derived literals can not be rebutted.

**Proposition** *3.1*
There exists no possible counter-argument for an argument structure $\langle \emptyset, q \rangle$.

**Proof:** Suppose that there exists a counter-argument $\langle \mathcal{A}, h \rangle$ for $\langle \emptyset, q \rangle$, then, $\mathcal{A} \cup \emptyset \cup \Pi$ should be a contradictory set, and therefore, $\mathcal{A} \cup \Pi$ should be a contradictory set. However, if $\mathcal{A} \cup \Pi$ is a contradictory, then $\langle \mathcal{A}, h \rangle$ is not an argument structure, because condition 2 of the definition of argument would not hold.  ∎

While the proposition above shows that an argument structure like $\langle \emptyset, q \rangle$ has no counter-arguments, the following proposition states that $\langle \emptyset, q \rangle$ can never be used as a counter-argument for any other argument structure $\langle \mathcal{A}, h \rangle$.

**Proposition** *3.2*
The argument structure $\langle \emptyset, q \rangle$ cannot be a counter-argument for any argument structure $\langle \mathcal{A}, h \rangle$.

**Proof:** Suppose that there exists $\langle \mathcal{A}, h \rangle$, such that $\langle \emptyset, q \rangle$ is a counter-argument for $\langle \mathcal{A}, h \rangle$. Then, $\mathcal{A} \cup \emptyset \cup \Pi$ is a contradictory set, and then $\mathcal{A} \cup \Pi$ is a contradictory set. However, if this is so, the set $\mathcal{A}$ could not be an argument, because condition 2 of the definition of argument would not hold. ∎

Given an argument structure $\langle \mathcal{A}, h \rangle$, there could be several counter-arguments attacking different points in $\mathcal{A}$, or different counter-arguments attacking the same point in $\mathcal{A}$. As we will show next, in order to verify whether an argument is non-defeated, all of its associated counter-arguments $\mathcal{B}_1, \mathcal{B}_2, \ldots, \mathcal{B}_k$ will be examined, each of them being a (defeasible) potential reason for rejecting $\mathcal{A}$. If any $\mathcal{B}_i$ is (somehow) "better" than, or unrelated to, $\mathcal{A}$, then $\mathcal{B}_i$ is a candidate for defeating $\mathcal{A}$. However, if the argument $\mathcal{A}$ is "better" than one $\mathcal{B}_i$ then $\mathcal{B}_i$ will not be taken in consideration as a defeater $\mathcal{A}$. This informal discussion shows the convenience of introducing a preference order among arguments.

### *3.2 Comparing Arguments*

The definition of a formal criterion for comparing arguments is a central problem in defeasible argumentation. Existing formalisms have adopted different solutions. Abstract argumentation systems usually assume an ordering in the set of all possible arguments (Dung, 1993b; Vreeswijk, 1997; Kowalski & Toni, 1996; Bondarenko *et al.*, 1997). For instance, in Dung's approach, an *argumentation framework* is a pair (*Args*, *attack*), where *Args* is the set of all possible arguments and *attack* is a binary relation on *Args*. If $(A, B) \in$ *attack* then argument $A$ attacks argument $B$.

In other formalisms, explicit priorities among program rules are given. Thus, the conflict between two rules can be solved. This approach is used in d-Prolog (Nute, 1988), Defeasible Logic (Nute, 1992), extensions of Defeasible Logic (Antoniou *et al.*, 2000a; Antoniou *et al.*, 1998), and logic programming without negation as failure (Kakas *et al.*, 1994; Dimopoulos & Kakas, 1995). In (Prakken & Sartor, 1997) it is also possible to reason (defeasibly) about priorities among rules.

An alternative is to use the specificity criterion (Poole, 1985), and no explicit order among rules or arguments need to be given. Finally, other formalisms use no comparison criterion (Gelfond & Lifschitz, 1990).

Next, we will introduce two criteria for comparing arguments. The first one based on specificity, and the second one based on priorities among program rules.

### *3.2.1 Generalized Specificity*

We will formally define a particular criterion called *generalized specificity* which allows to discriminate between two conflicting arguments. Intuitively, this notion of specificity favors two aspects in an argument: it prefers an argument (1) with greater information content or (2) with less use of rules (more direct.) In other words, an argument will be deemed better than another if it is *more precise* or *more concise* (see the Example 3.5 below.)

The next definition characterizes the specificity criterion, defined first in (Poole, 1985) and extended later to be used in defeasible argumentation (Simari & Loui, 1992; Simari *et al.*, 1994b). Here, it is adapted to fit in the DeLP framework. As stated before, $\mathcal{P} \mathrel{|\!\sim} h$ means that $h$ has a defeasible derivation from $\mathcal{P}$, and $\mathcal{P} \vdash h$ means that $h$ has a strict derivation from $\mathcal{P}$.

**Definition** *3.5* (*Specificity*)
Let $\mathcal{P}=(\Pi, \Delta)$ be a *de.l.p.*, and let $\Pi_G$ be the set of all strict rules from $\Pi$ (without including facts.) Let $\mathcal{F}$ be the set of all literals that have a defeasible derivation from $\mathcal{P}$ ($\mathcal{F}$ will be considered as a set of facts.) Let $\langle \mathcal{A}_1, h_1 \rangle$ and $\langle \mathcal{A}_2, h_2 \rangle$ be two argument structures obtained from $\mathcal{P}$. $\langle \mathcal{A}_1, h_1 \rangle$ is *strictly more specific than* $\langle \mathcal{A}_2, h_2 \rangle$ (denoted $\langle \mathcal{A}_1, h_1 \rangle \succ \langle \mathcal{A}_2, h_2 \rangle$) if the following conditions hold:

1. For all $H \subseteq \mathcal{F}$: if $\Pi_G \cup H \cup \mathcal{A}_1 \mathrel{|\!\sim} h_1$ and $\Pi_G \cup H \not\vdash h_1$,
   then $\Pi_G \cup H \cup \mathcal{A}_2 \mathrel{|\!\sim} h_2$, and
2. there exists $H' \subseteq \mathcal{F}$ such that $\Pi_G \cup H' \cup \mathcal{A}_2 \mathrel{|\!\sim} h_2$
   and $\Pi_G \cup H' \not\vdash h_2$, and $\Pi_G \cup H' \cup \mathcal{A}_1 \mathrel{|\!\not\sim} h_1$

As mentioned before, it is not possible to have a defeasible derivation for a literal from a set of rules without facts. Therefore, from the set $\Pi_G \cup \mathcal{A}_1$ it is not possible to have a defeasible derivation for $h_1$. However, from $\Pi_G \cup H \cup \mathcal{A}_1$ it could be possible because $H$ is a set of literals (facts.) Thus, when $\Pi_G \cup H \cup \mathcal{A}_1 \mathrel{|\!\sim} h_1$ holds, we say that the set $H$ *activates* $\langle \mathcal{A}_1, h_1 \rangle$, or $H$ is an *activation set* of $\langle \mathcal{A}_1, h_1 \rangle$.

**Example** *3.5*
Consider the program $\mathcal{P}_{2.1}$ of Example 2.1, the argument structure

$$\langle \mathcal{A}_1, h_1 \rangle = \langle \{\sim\!\mathit{flies}(\mathit{tina}) \prec \mathit{chicken}(\mathit{tina})\}, \sim\!\mathit{flies}(\mathit{tina}) \rangle$$

is strictly more specific than $\langle \mathcal{A}_2, h_2 \rangle = \langle \{\mathit{flies}(\mathit{tina}) \prec \mathit{bird}(\mathit{tina})\}, \mathit{flies}(\mathit{tina}) \rangle$ because $\langle \mathcal{A}_1, h_1 \rangle$ is 'more direct' than $\langle \mathcal{A}_2, h_2 \rangle$ (observe that $\langle \mathcal{A}_2, h_2 \rangle$ uses the strict rule $\mathit{bird}(\mathit{tina}) \leftarrow \mathit{chicken}(\mathit{tina})$.) In this example every activation set $H$ of $\langle \mathcal{A}_1, h_1 \rangle$ also activates $\langle \mathcal{A}_2, h_2 \rangle$, but the set $H' = \{\mathit{bird}(\mathit{tina})\}$ activates $\langle \mathcal{A}_2, h_2 \rangle$ but does not activate $\langle \mathcal{A}_1, h_1 \rangle$. Using the same program, the argument structure

$$\langle \mathcal{A}_3, h_3 \rangle = \langle \{\mathit{flies}(\mathit{tina}) \prec \mathit{chicken}(\mathit{tina}), \mathit{scared}(\mathit{tina})\}, \mathit{flies}(\mathit{tina}) \rangle$$

will be preferred to $\langle \mathcal{A}_1, h_1 \rangle$, because $\langle \mathcal{A}_3, h_3 \rangle$ is based in more information (the literals $\mathit{chicken}(\mathit{tina})$ and $\mathit{scared}(\mathit{tina})$) than $\langle \mathcal{A}_1, h_1 \rangle$. Observe that every activation set $H$ of $\langle \mathcal{A}_3, h_3 \rangle$ contains the literals $\mathit{chicken}(\mathit{tina})$ and $\mathit{scared}(\mathit{tina})$, thus $H$ actives $\langle \mathcal{A}_1, h_1 \rangle$. However, the set $H' = \{\mathit{chicken}(\mathit{tina})\}$ activates $\langle \mathcal{A}_1, h_1 \rangle$, and $H'$ does not activate $\langle \mathcal{A}_3, h_3 \rangle$. Therefore, $\langle \mathcal{A}_3, h_3 \rangle$ is strictly more specific than $\langle \mathcal{A}_1, h_1 \rangle$. ∎

**Definition** *3.6* (*Equi-Specificity*)
Two arguments $\langle \mathcal{A}_1, h_1 \rangle$ and $\langle \mathcal{A}_2, h_2 \rangle$ are equi-specific, denoted $\langle \mathcal{A}_1, h_1 \rangle \equiv \langle \mathcal{A}_2, h_2 \rangle$, iff $\mathcal{A}_1 = \mathcal{A}_2$, and the literal $h_2$ has a strict derivation from $\Pi \cup \{h_1\}$, and the literal $h_1$ has a strict derivation from $\Pi \cup \{h_2\}$.

**Lemma** *3.1*
Equi-specificity is an equivalence relation.

**Proof:** Trivial by definition. ∎

The following proposition shows that strictly derived literals are 'equi-specific'.

**Proposition** *3.3*
If $\mathcal{A}_1 = \emptyset$ and $\mathcal{A}_2 = \emptyset$, then $\langle \mathcal{A}_1, h_1 \rangle \equiv \langle \mathcal{A}_2, h_2 \rangle$

**Proof:** If $\mathcal{A}_1$ and $\mathcal{A}_2$ are empty, then $\mathcal{A}_1 = \mathcal{A}_2$, $\Pi \vdash h_1$ and $\Pi \vdash h_2$. Hence, $\Pi \cup \{h_1\} \vdash h_2$ and $\Pi \cup \{h_2\} \vdash h_1$. Observe also that since $\Pi$ is not contradictory, then $\Pi \cup \{h_1, h_2\}$ is not contradictory. ∎

**Proposition** *3.4*
If two argument structures $\langle \mathcal{A}_1, h_1 \rangle$ and $\langle \mathcal{A}_2, h_2 \rangle$ are equi-specific then $h_1$ and $h_2$ cannot disagree.

**Proof:** If $\langle \mathcal{A}_1, h_1 \rangle \equiv \langle \mathcal{A}_2, h_2 \rangle$, then $\mathcal{A}_1 = \mathcal{A}_2$, and $\Pi \cup \{h_1\} \vdash h_2$, and $\Pi \cup \{h_2\} \vdash h_1$. Suppose that $h_1$ and $h_2$ disagree, then the set $\Pi \cup \{h_1, h_2\}$ is contradictory. Since $\Pi \cup \{h_1\} \vdash h_2$, then $\Pi \cup \{h_1\}$ is contradictory, and therefore, $\langle \mathcal{A}_1, h_1 \rangle$ cannot be an argument structure because it does not satisfy condition 2 of argument definition. ∎

### 3.2.2 Argument comparison using rule's priorities

Some formalisms define explicit priorities among rules and use these priorities for deciding between competing conclusions. The use of these priorities is usually embedded in the derivation mechanism and competing rules are compared individually during the derivation process. In such formalisms the derivation notion is bound to one single comparison criterion.

In DeLP in order to decide between competing conclusions the arguments that support the conclusions are compared. Thus, the comparison criterion is independent of the derivation process, and could be replaced in a modular way. Next, we will introduce a particular comparison criterion which uses a form of ordering the rules by their priority as an example.

We will show how a comparison criterion between arguments based on rule priorities can be formulated. We will assume that explicit priorities among defeasible rules are given with the program. Since strict rules represent sound information, there will be no priorities among them. Priorities will be allowed only between two defeasible rules. As showed in Proposition 3.1, a literal that has a strict derivation has no counter-argument. Therefore, implicitly, a strict derivation will be preferred over other arguments that use defeasible rules. Many comparison criteria could be defined. The priority based criterion defined below is but one example.

**Definition** *3.7*
Let $\mathcal{P}$ be *de.l.p.* and ">" a preference relation explicitly defined among defeasible rules. Given two argument structures $\langle \mathcal{A}_1, h_1 \rangle$ and $\langle \mathcal{A}_2, h_2 \rangle$, the argument $\langle \mathcal{A}_1, h_1 \rangle$ will be preferred over $\langle \mathcal{A}_2, h_2 \rangle$ if:

1. there exists at least one rule $r_a \in \mathcal{A}_1$, and one rule $r_b \in \mathcal{A}_2$, such that $r_a > r_b$,
2. and there is no $r'_b \in \mathcal{A}_2$, and $r'_a \in \mathcal{A}_1$, such that $r'_b > r'_a$.

**Example** *3.6*
*Consider the following de.l.p.*

$$\mathcal{P}_{3.6} = \left\{ \begin{array}{l} buy\_stock(T) \prec good\_price(T) \\ \sim buy\_stock(T) \prec risky\_company(T) \\ risky\_company(T) \prec in\_fusion(T, Y) \\ good\_price(acme) \\ in\_fusion(acme, steel) \end{array} \right\}$$

*And the priority:*

$$buy\_stock(T) \prec good\_price(T) \quad < \quad \sim buy\_stock(T) \prec risky\_company(T)$$

*Using the criterion of Definition 3.7, the argument structure*

$$\left\langle \left\{ \begin{array}{l} \sim buy\_stock(acme) \prec risky\_company(acme) \\ risky\_company(acme) \prec in\_fusion(acme, steel) \end{array} \right\}, \sim buy\_stock(acme) \right\rangle$$

*will be preferred over* $\langle \{buy\_stock(acme) \prec good\_price(acme)\}, buy\_stock(acme) \rangle$
∎

A more sophisticated criterion could be obtained combining the two defined above. For example, considering first generalized specificity, and if no argument is preferred, then use the existing priorities.

## 4 Defeaters and Argumentation Lines

Given an argument structure $\langle \mathcal{A}_1, h_1 \rangle$, and a counter-argument $\langle \mathcal{A}_2, h_2 \rangle$ for $\langle \mathcal{A}_1, h_1 \rangle$, these two arguments can be compared in order to decide which one prevails. Namely, if the counter-argument $\langle \mathcal{A}_2, h_2 \rangle$ is better than $\langle \mathcal{A}_1, h_1 \rangle$ w.r.t the comparison criterion used, then $\langle \mathcal{A}_2, h_2 \rangle$ will be called a *proper defeater* for $\mathcal{A}_1$. If neither argument is better, nor worse, than the other, a blocking situation occurs, and we will say that $\langle \mathcal{A}_2, h_2 \rangle$ is a *blocking defeater* for $\langle \mathcal{A}_1, h_1 \rangle$. If $\langle \mathcal{A}_1, h_1 \rangle$ is better than $\langle \mathcal{A}_2, h_2 \rangle$, then $\langle \mathcal{A}_2, h_2 \rangle$ will not be considered as a defeater for $\langle \mathcal{A}_1, h_1 \rangle$.

Although a preference criterion is required for comparing arguments, the notion of *defeating argument* can be formulated independently of the particular argument-discriminating criterion that is being used. From now on, we will abstract away from the comparison criterion, assuming there exists a comparison criterion among arguments that we will denote "$\succ$". For the examples in the rest of the paper we will assume that "$\succ$" means "strictly more specific" as defined above.

**Definition** *4.1* (*Proper Defeater*)
Let $\langle \mathcal{A}_1, h_1 \rangle$ and $\langle \mathcal{A}_2, h_2 \rangle$ be two argument structures. $\langle \mathcal{A}_1, h_1 \rangle$ is a *proper defeater* for $\langle \mathcal{A}_2, h_2 \rangle$ at literal $h$, if and only if there exists a sub-argument $\langle \mathcal{A}, h \rangle$ of $\langle \mathcal{A}_2, h_2 \rangle$ such that $\langle \mathcal{A}_1, h_1 \rangle$ counter-argues $\langle \mathcal{A}_2, h_2 \rangle$ at $h$, and $\langle \mathcal{A}_1, h_1 \rangle \succ \langle \mathcal{A}, h \rangle$.

Observe that in the previous definition, the argument structure $\langle \mathcal{A}_1, h_1 \rangle$ is compared with the disagreement subargument $\langle \mathcal{A}, h \rangle$.

Defeasible Logic Programming    An Argumentative Approach    17

**Example** *4.1*
In Example 3.5, since $\langle \mathcal{A}_1, h_1 \rangle$ is a counter argument for $\langle \mathcal{A}_2, h_2 \rangle$, and $\langle \mathcal{A}_1, h_1 \rangle \succ \langle \mathcal{A}_2, h_2 \rangle$, then $\langle \mathcal{A}_1, h_1 \rangle$ is a proper defeater for $\langle \mathcal{A}_2, h_2 \rangle$. Observe that $\langle \mathcal{A}_3, h_3 \rangle$ is a proper defeater for $\langle \mathcal{A}_1, h_1 \rangle$. ∎

**Definition** *4.2* (*Blocking Defeater*)
Let $\langle \mathcal{A}_1, h_1 \rangle$ and $\langle \mathcal{A}_2, h_2 \rangle$ be two argument structures. $\langle \mathcal{A}_1, h_1 \rangle$ is a *blocking defeater* for $\langle \mathcal{A}_2, h_2 \rangle$ at literal $h$, if and only if there exists a sub-argument $\langle \mathcal{A}, h \rangle$ of $\langle \mathcal{A}_2, h_2 \rangle$ such that $\langle \mathcal{A}_1, h_1 \rangle$ counter-argues $\langle \mathcal{A}_2, h_2 \rangle$ at $h$, and $\langle \mathcal{A}_1, h_1 \rangle$ is unrelated by the preference order to $\langle \mathcal{A}, h \rangle$, i.e., $\langle \mathcal{A}_1, h_1 \rangle \not\succ \langle \mathcal{A}, h \rangle$, and $\langle \mathcal{A}, h \rangle \not\succ \langle \mathcal{A}_1, h_1 \rangle$.

**Example** *4.2*
*The Nixon Diamond provides the proverbial example of blocking defeaters. Consider the de.l.p. $\mathcal{P}_{2.2}$ of Example 2.2. From $\mathcal{P}_{2.2}$, the following argument structures can be obtained:*
$\langle \mathcal{A}_1, h_1 \rangle = \langle \{pacifist(nixon) \prec quaker(nixon)\}, pacifist(nixon) \rangle$ and
$\langle \mathcal{A}_2, h_2 \rangle = \langle \{\sim pacifist(nixon) \prec republican(nixon)\}, \sim pacifist(nixon) \rangle$.

*The argument $\langle \mathcal{A}_2, h_2 \rangle$ is a blocking defeater for $\langle \mathcal{A}_1, h_1 \rangle$, and vice versa. As it will be shown below, in DeLP the answer for the query 'pacifist(nixon)' will be undecided.* ∎

**Definition** *4.3* (*Defeater*)
The argument structure $\langle \mathcal{A}_1, h_1 \rangle$ is a *defeater* for $\langle \mathcal{A}_2, h_2 \rangle$, if and only if either:

1. $\langle \mathcal{A}_1, h_1 \rangle$ is a *proper* defeater for $\langle \mathcal{A}_2, h_2 \rangle$; or
2. $\langle \mathcal{A}_1, h_1 \rangle$ is a *blocking* defeater for $\langle \mathcal{A}_2, h_2 \rangle$.

Thus, a defeater for an argument structure can be identified as proper or blocking. As we will show below, this distinction will be considered by the warrant procedure. It is interesting to note, that most argumentation formalisms make no distinction between proper or blocking defeaters, and some of them only consider proper defeaters. The following proposition shows that during the argumentation process it is not possible to attack a subargument $\langle \mathcal{A}, h \rangle$ with an argument $\langle \mathcal{A}_1, h_1 \rangle$ that is equi-specific to $\langle \mathcal{A}, h \rangle$.

**Proposition** *4.1*
If $\langle \mathcal{A}_1, h_1 \rangle$ is a defeater (proper or blocking) for $\langle \mathcal{A}_2, h_2 \rangle$, and $\langle \mathcal{A}, h \rangle$ is the corresponding disagreement subargument, then it cannot be the case that $\langle \mathcal{A}_1, h_1 \rangle \equiv \langle \mathcal{A}, h \rangle$.

**Proof:** Proposition 3.4 shows that if $\langle \mathcal{A}_1, h_1 \rangle \equiv \langle \mathcal{A}, h \rangle$, then $h_1$ and $h$ could not disagree. Thus, $\langle \mathcal{A}_1, h_1 \rangle$ cannot be a counter-argument for $\langle \mathcal{A}_2, h_2 \rangle$ at $h$. ∎

In order to establish whether an argument structure $\langle \mathcal{A}_0, h_0 \rangle$ is non-defeated, all defeaters for $\langle \mathcal{A}_0, h_0 \rangle$ have to be considered. Suppose that $\langle \mathcal{A}_1, h_1 \rangle$ is a defeater for $\langle \mathcal{A}_0, h_0 \rangle$, since $\langle \mathcal{A}_1, h_1 \rangle$ is an argument structure, then defeaters for $\langle \mathcal{A}_1, h_1 \rangle$ may exist, and so on. In this manner, a sequence of argument structures is created, where each element of the sequence defeats its predecessor. We formalize this notion next.

**Definition** *4.4* (*Argumentation Line*)
Let $\mathcal{P}$ a *de.l.p.* and $\langle \mathcal{A}_0, h_0 \rangle$ an argument structure obtained from $\mathcal{P}$. An *argumentation line* for $\langle \mathcal{A}_0, h_0 \rangle$ is a sequence of argument structures from $\mathcal{P}$, denoted $\Lambda = [\langle \mathcal{A}_0, h_0 \rangle, \langle \mathcal{A}_1, h_1 \rangle, \langle \mathcal{A}_2, h_2 \rangle \langle \mathcal{A}_3, h_3 \rangle, \ldots]$, where each element of the sequence $\langle \mathcal{A}_i, h_i \rangle$, $i > 0$, is a defeater of its predecessor $\langle \mathcal{A}_{i-1}, h_{i-1} \rangle$.

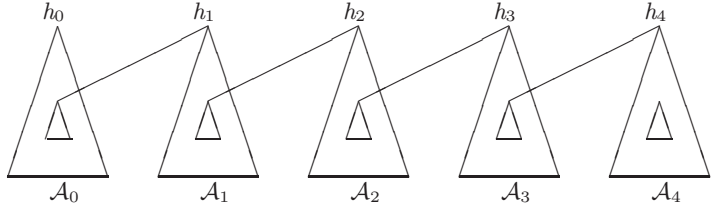

Fig. 3. Argumentation line

As defined above, an argumentation line could result in an infinite sequence of arguments. However, in the following section we will impose some restrictions over the argumentation lines and only finite sequences will be allowed.

In each argumentation line $\Lambda = [\langle \mathcal{A}_0, h_0 \rangle, \langle \mathcal{A}_1, h_1 \rangle, \langle \mathcal{A}_2, h_2 \rangle \langle \mathcal{A}_3, h_3 \rangle, \ldots]$, the argument $\langle \mathcal{A}_0, h_0 \rangle$ is supporting the main query $h_0$, and every argument $\langle \mathcal{A}_i, h_i \rangle$ defeats its predecessor $\langle \mathcal{A}_{i-1}, h_{i-1} \rangle$. Then, $\langle \mathcal{A}_0, h_0 \rangle$ becomes a *supporting* argument for $h_0$, $\langle \mathcal{A}_1, h_1 \rangle$ an *interfering* argument, $\langle \mathcal{A}_2, h_2 \rangle$ a supporting argument, $\langle \mathcal{A}_3, h_3 \rangle$ an interfering one, and so on. Thus, an argumentation line can be split in two disjoint sets: $\Lambda_S$ of supporting arguments, and $\Lambda_I$ of interfering arguments.

**Definition** *4.5* (*Supporting and Interfering argument structures*)
Let $\Lambda = [\langle \mathcal{A}_0, h_0 \rangle, \langle \mathcal{A}_1, h_1 \rangle, \langle \mathcal{A}_2, h_2 \rangle \langle \mathcal{A}_3, h_3 \rangle, \ldots]$ an argumentation line, we define the set of supporting argument structures $\Lambda_S = \{ \langle \mathcal{A}_0, h_0 \rangle, \langle \mathcal{A}_2, h_2 \rangle, \langle \mathcal{A}_4, h_4 \rangle, \ldots \}$, and the set of interfering argument structures $\Lambda_I = \{ \langle \mathcal{A}_1, h_1 \rangle, \langle \mathcal{A}_3, h_3 \rangle, \ldots \}$.

Given an argument structure $\langle \mathcal{A}_0, h_0 \rangle$, there can be many defeaters for $\langle \mathcal{A}_0, h_0 \rangle$, and each of them will generate a different argumentation line. Observe also, that in these argumentation lines any of the arguments could have more than one defeater generating more argumentation lines starting with $\langle \mathcal{A}_0, h_0 \rangle$.

**Example** *4.3*
Consider a program $\mathcal{P}$ where $\langle \mathcal{A}_1, h_1 \rangle$ defeats $\langle \mathcal{A}_0, h_0 \rangle$, and $\langle \mathcal{A}_2, h_2 \rangle$ also defeats $\langle \mathcal{A}_0, h_0 \rangle$. Up to this point there are two argumentation lines. Now suppose that $\langle \mathcal{A}_3, h_3 \rangle$ defeats $\langle \mathcal{A}_1, h_1 \rangle$, and that both $\langle \mathcal{A}_4, h_4 \rangle$ and $\langle \mathcal{A}_5, h_5 \rangle$ defeat $\langle \mathcal{A}_2, h_2 \rangle$, then there are several argumentation lines starting with $\langle \mathcal{A}_0, h_0 \rangle$, here we show three of them:

$$\Lambda_1 = [\langle \mathcal{A}_0, h_0 \rangle, \langle \mathcal{A}_1, h_1 \rangle, \langle \mathcal{A}_3, h_3 \rangle]$$
$$\Lambda_2 = [\langle \mathcal{A}_0, h_0 \rangle, \langle \mathcal{A}_2, h_2 \rangle, \langle \mathcal{A}_4, h_4 \rangle]$$
$$\Lambda_3 = [\langle \mathcal{A}_0, h_0 \rangle, \langle \mathcal{A}_2, h_2 \rangle, \langle \mathcal{A}_5, h_5 \rangle]$$

■

Therefore, a process that considers all possible argumentation lines is needed. Before defining such a process, we will introduce some restrictions over argumentation lines.

### *4.1 Acceptable Argumentation Lines*

In this section we will show several undesirable situations that may arise in an argumentation line leading to an infinite sequence of defeaters. We will then impose certain constraints over the argumentation lines in order to avoid these problematic situations. Some of these situations were reported first in (Simari *et al.*, 1994b).

In the related literature, an argument structure $\langle \mathcal{A}, h \rangle$ is said to be "self-defeating" if $\langle \mathcal{A}, h \rangle$ is a defeater for itself. If $\langle \mathcal{A}, h \rangle$ is a self-defeating argument structure then an argumentation line starting with $\langle \mathcal{A}, h \rangle$ will be infinite (see Figure 4.)

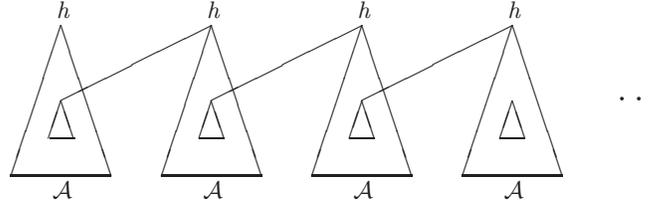

Fig. 4. Infinite argumentation line with a self defeating argument

Many approaches of defeasible argumentation have to deal with self-defeating arguments. As stated next, arguments in DeLP will never be self-defeating.

**Proposition** *4.2*
In DeLP no argument structure can be self-defeating.

**Proof:** Assume that $\langle \mathcal{A}, h \rangle$ is self-defeating, then $\langle \mathcal{A}, h \rangle$ would be a defeater for itself, and there should exist a counter-argument point $q$ in $\mathcal{A}$ such that $\Pi \cup \{h, q\}$ is contradictory. Therefore $\Pi \cup \mathcal{A}$ would be contradictory, and $\langle \mathcal{A}, h \rangle$ would not be an argument structure.  ∎

Another problematic situation, mentioned by Henry Prakken in (Prakken & Vreeswijk, 2000), are reciprocal defeaters. This happens when a pair of arguments defeat each other. Example 4.4 and Figure 5 shows that case: $\langle \mathcal{A}_2, d \rangle$ defeats $\langle \mathcal{A}_1, b \rangle$, attacking the subargument $\langle \mathcal{B}, \sim d \rangle$, but $\langle \mathcal{A}_1, b \rangle$ also defeats $\langle \mathcal{A}_2, d \rangle$ attacking the subargument $\langle \mathcal{A}, \sim b \rangle$.

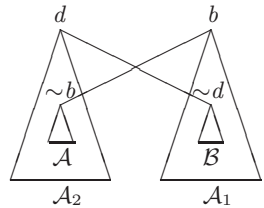

Fig. 5. Reciprocal defeaters

**Example** *4.4*
Consider the following *de.l.p.*: $\{\ (d \prec \sim b, c),\ (b \prec \sim d, a),\ (\sim b \prec a),\ (\sim d \prec c),\ (a), (c)\ \}$. The argument $\langle \mathcal{A}_2, d \rangle = \langle \{(d \prec \sim b, c), (\sim b \prec a)\}, d \rangle$ is a proper defeater for $\langle \mathcal{A}_1, b \rangle = \langle \{(b \prec \sim d, a), (\sim d \prec c)\}, b \rangle$, and vice versa. Observe that $\langle \mathcal{A}_2, d \rangle$ is *strictly more specific than the subargument* $\langle\{\sim d \prec c\},\sim d\rangle$ *of* $\langle\mathcal{A}_1, b\rangle$, *and* $\langle\mathcal{A}_1, b\rangle$ *is strictly more specific than* $\langle\{\sim b \prec a\},\sim b\rangle$. ∎

Clearly, this situation is undesirable as it leads to the construction of an infinite sequence of arguments. Therefore, reciprocal defeaters must be detected and avoided. The analysis prompted by Henry Prakken's remark led to investigate other types of undesirable situations in argumentation.

A circular argumentation is obtained when an argument structure is reintroduced again in an argumentation line to defend itself. Figure 6 shows an example of circular argumentation. There, the same argument $\mathcal{A}$ is reintroduced down the line as a supporting argument for itself leading to an infinite argumentation line. Circular argumentation was discussed first in (Simari et al., 1994b) as a particular case of *fallacious argumentation*.

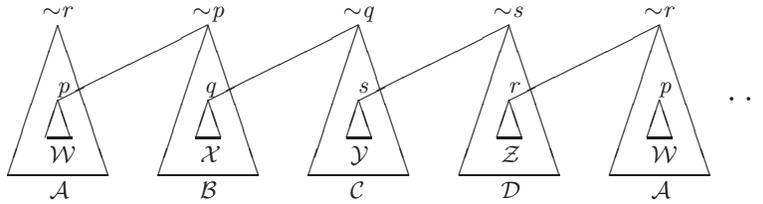

Fig. 6. Circular argumentation

In order to avoid circular argumentation we need to impose the condition that *no argument can be reintroduced in the same argumentation line*. However, a more subtle case of circular argumentation happens with the reintroduction of a subargument. Figure 7 shows this situation: argument $\mathcal{B}$ is a defeater for $\mathcal{A}$, and $\mathcal{W}$ is the disagreement sub-argument. Later in the line, argument $\mathcal{W}$ could be reintroduced as a defeater, allowing the reintroduction of $\mathcal{B}$. Although the cycle can be detected and broken when $\mathcal{B}$ is reintroduced, the fallacious situation is the reintroduction of a subargument that was defeated earlier in the line.

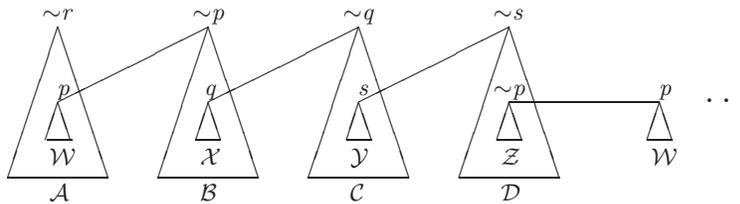

Fig. 7. Circular argumentation with a sub-argument

A different, but also undesirable, situation is shown in Figure 8. There, the same argument $\mathcal{A}$ becomes *both* a supporting and an interfering argument of itself. This situation arises because the supporting argument $\mathcal{C}$ has a subargument $\mathcal{Z}$ for the literal $r$, which is contradictory with arguing in favor of $\sim r$ (argument $\mathcal{A}$.) The introduction of an argument like $\mathcal{C}$ should be avoided in a sound argumentation line. Clearly, there should be agreement among supporting arguments (respectively interfering) in any argumentation line. This is expressed formally with the notion of argument *concordance* as proposed in (Simari et al., 1994b) and recalled next.

**Definition** *4.6* (*Concordance*)
Let $\mathcal{P}=(\Pi, \Delta)$ be a *de.l.p.*. Two arguments $\langle \mathcal{A}_1, h_1 \rangle$ and $\langle \mathcal{A}_2, h_2 \rangle$ are *concordant* iff the set $\Pi \cup \mathcal{A}_1 \cup \mathcal{A}_2$ is non-contradictory. More generally, a set of argument structures $\{\langle \mathcal{A}_i, h_i \rangle\}_{i=1}^{n}$ is concordant iff $\Pi \cup \bigcup_{i=1}^{n} \mathcal{A}_i$ is non-contradictory.

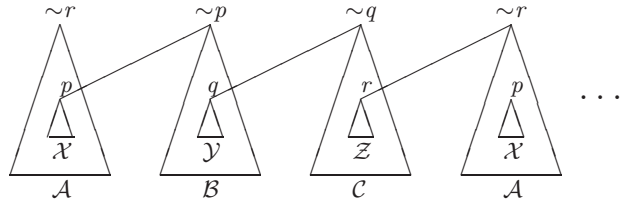

Fig. 8. Contradictory argumentation line

In the case shown in Figure 8, the cycle could be detected and broken disallowing the reintroduction of argument $\mathcal{A}$. However, the fallacious move is the use of argument $\mathcal{C}$ that makes the set of supporting arguments non-concordant. Observe that the status of the first argument of the line will change depending on which criterion we use: on one hand, if we allow the use of $\mathcal{C}$, and just forbid the reintroduction of $\mathcal{A}$, the first argument in the line would not be defeated; on the other hand, if $\mathcal{C}$ is forbidden, the first argument of the line will be defeated.

Therefore, we will establish the condition that *the set of supporting arguments of an argumentation line must be concordant, and the same must hold for the set of interfering arguments*. Thus, the introduction of argument $\mathcal{C}$ in the example of Figure 8 will not be allowed.

A different ill-formed situation corresponds to the use of a blocking defeater to defeat a blocking defeater. Consider the following *de.l.p.*:

$$\left\{ \begin{array}{ll} dangerous(X) \prec tiger(X) & tiger(hobbes) \\ \sim dangerous(X) \prec baby(X) & baby(hobbes) \\ \sim dangerous(X) \prec pet(X) & pet(hobbes) \end{array} \right\}$$

Here, $\mathcal{A}_1 = \{ \sim dangerous(hobbes) \prec baby(hobbes) \}$ supports $\sim dangerous(hobbes)$. The argument $\mathcal{A}_2 = \{dangerous(hobbes) \prec tiger(hobbes)\}$ is a blocking defeater for the argument $\mathcal{A}_1$, and $\mathcal{A}_3 = \{\sim dangerous(hobbes) \prec pet(hobbes)\}$ is a blocking defeater for $\mathcal{A}_2$. The following argumentation line may be obtained: $[\mathcal{A}_1, \mathcal{A}_2, \mathcal{A}_3]$. Observe that although $\mathcal{A}_2$ is a defeater for $\mathcal{A}_3$, $\mathcal{A}_2$ is not introduced again because it was already used in the line.

If the argumentation line $[\mathcal{A}_1, \mathcal{A}_2, \mathcal{A}_3]$ is accepted, then $\mathcal{A}_3$ defeats $\mathcal{A}_2$, reinstating $\mathcal{A}_1$. However, a blocking argument $\mathcal{A}_3$ is being used for defeating a blocking defeater $\mathcal{A}_2$, but $\mathcal{A}_2$ was already blocked by $\mathcal{A}_1$. This is equivalent to accepting that in a blocking situation having two arguments for "$\sim dangerous(hobbes)$" is preferred over having just one argument for the contrary. In order to avert this problem, when an argument $\langle \mathcal{A}_i, h_i \rangle$ is used as a blocking defeater for $\langle \mathcal{A}_{i-1}, h_{i-1} \rangle$ during the construction of an argumentation line, only a proper defeater could be used for defeating $\langle \mathcal{A}_i, h_i \rangle$.

In DeLP, the undesirable situations mentioned above are avoided by requiring all argumentation lines to be *acceptable* as defined next.

**Definition** *4.7* (*Acceptable argumentation line*)
Let $\Lambda = [\langle \mathcal{A}_1, h_1 \rangle, \ldots, \langle \mathcal{A}_i, h_i \rangle, \ldots, \langle \mathcal{A}_n, h_n \rangle]$ be an argumentation line. $\Lambda$ is an *acceptable argumentation line* iff:

1. $\Lambda$ is a finite sequence.
2. The set $\Lambda_S$, of supporting arguments is concordant, and the set $\Lambda_I$ of interfering arguments is concordant.
3. No argument $\langle \mathcal{A}_k, h_k \rangle$ in $\Lambda$ is a subargument of an argument $\langle \mathcal{A}_i, h_i \rangle$ appearing earlier in $\Lambda$ ($i < k$.)
4. For all $i$, such that the argument $\langle \mathcal{A}_i, h_i \rangle$ is a blocking defeater for $\langle \mathcal{A}_{i-1}, h_{i-1} \rangle$, if $\langle \mathcal{A}_{i+1}, h_{i+1} \rangle$ exists, then $\langle \mathcal{A}_{i+1}, h_{i+1} \rangle$ is a proper defeater for $\langle \mathcal{A}_i, h_i \rangle$.

It is interesting to note that changes in the definition of acceptable argumentation line may produce a different behavior of the formalism. Thus, this definition could be used as a way of tuning the system to obtain different results.

## 5 Warrant through Dialectical Analysis

In DeLP a literal $h$ will be warranted[2] if there exists a non-defeated argument structure $\langle \mathcal{A}, h \rangle$. In order to establish whether $\langle \mathcal{A}, h \rangle$ is non-defeated, the set of defeaters for $\mathcal{A}$ will be considered. Since each defeater $\mathcal{D}$ for $\mathcal{A}$ is itself an argument structure, defeaters for $\mathcal{D}$ will in turn be considered, and so on. Therefore, as stated in Example 4.3, more than one argumentation line could arise, leading to a tree structure that we will call dialectical tree.

**Definition** *5.1* (*Dialectical Tree*)
Let $\langle \mathcal{A}_0, h_0 \rangle$ be an argument structure from a program $\mathcal{P}$. A dialectical tree for $\langle \mathcal{A}_0, h_0 \rangle$, denoted $\mathcal{T}_{\langle \mathcal{A}_0, h_0 \rangle}$, is defined as follows:

1. The root of the tree is labeled with $\langle \mathcal{A}_0, h_0 \rangle$.
2. Let $N$ be a non-root node of the tree labeled $\langle \mathcal{A}_n, h_n \rangle$, and
   $\Lambda = [\langle \mathcal{A}_0, h_0 \rangle, \langle \mathcal{A}_1, h_1 \rangle, \langle \mathcal{A}_2, h_2 \rangle, \ldots, \langle \mathcal{A}_n, h_n \rangle]$ the sequence of labels of the path from the root to $N$. Let $\langle \mathcal{B}_1, q_1 \rangle$, $\langle \mathcal{B}_2, q_2 \rangle$, …, $\langle \mathcal{B}_k, q_k \rangle$ be all the defeaters for $\langle \mathcal{A}_n, h_n \rangle$.
   For each defeater $\langle \mathcal{B}_i, q_i \rangle$ ($1 \leq i \leq k$), such that, the argumentation line $\Lambda' = [\langle \mathcal{A}_0, h_0 \rangle, \langle \mathcal{A}_1, h_1 \rangle, \langle \mathcal{A}_2, h_2 \rangle, \ldots, \langle \mathcal{A}_n, h_n \rangle, \langle \mathcal{B}_i, q_i \rangle]$ is acceptable, then the node $N$ has a child $N_i$ labeled $\langle \mathcal{B}_i, q_i \rangle$.
   If there is no defeater for $\langle \mathcal{A}_n, h_n \rangle$ or there is no $\langle \mathcal{B}_i, q_i \rangle$ such that $\Lambda'$ is acceptable, then $N$ is a leaf.

---

[2] In previous work we have used the term *justification*. We decide to adopt the term *warrant* in order to unify the terminology with other approaches.

In a dialectical tree every node (except the root) represents a defeater (proper or blocking) of its parent, and leaves correspond to non-defeated arguments. Each path from the root to a leaf corresponds to one different acceptable argumentation line. As we will show in Example 5.1 the dialectical tree provides a structure for considering all the possible acceptable argumentation lines that can be generated for deciding whether an argument is defeated. We call this tree *dialectical* because it represents an exhaustive dialectical analysis for the argument in its root.

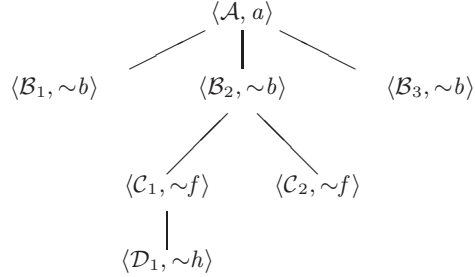

Fig. 9. Dialectical tree for Example 5.1

**Example** *5.1*
*Consider the following de.l.p.:*

$$\left\{ \begin{array}{llll} a \prec b & \sim b \prec e & \sim b \prec c, f & \sim f \prec i \\ b \prec c & e & f \prec g & i \\ c & \sim f \prec g, h & g & \sim h \prec k \\ \sim b \prec c, d & h \prec j & k & \\ d & j & & \end{array} \right\}$$

Here, the literal $a$ is supported by $\langle \mathcal{A}, a \rangle = \langle \{(a \prec b), (b \prec c)\}, a \rangle$ and there exist three defeaters for it, each of them starting three different argumentation lines: $\langle \mathcal{B}_1, \sim b \rangle = \langle \{(\sim b \prec c, d)\}, \sim b \rangle$, $\langle \mathcal{B}_2, \sim b \rangle = \langle \{(\sim b \prec c, f), (f \prec g)\}, \sim b \rangle$, and $\langle \mathcal{B}_3, \sim b \rangle = \langle \{(\sim b \prec e)\}, \sim b \rangle$. The first two are proper defeaters and the last one is a blocking defeater. Observe that the argument structure $\langle \mathcal{B}_1, \sim b \rangle$ has the counter-argument $\langle \{b \prec c\}, b \rangle$, but it is not a defeater because the former is more specific. Thus, no defeaters for $\langle \mathcal{B}_1, \sim b \rangle$ exist and the argumentation line ends there.
The argument structure $\langle \mathcal{B}_3, \sim b \rangle$ has a blocking defeater: $\langle \{b \prec c\}, b \rangle$. Note that $\langle \{b \prec c\}, b \rangle$ is the disagreement subargument of $\langle \mathcal{A}, a \rangle$, therefore, it cannot be introduced because it produces an argumentation line that is not acceptable.
The argument structure $\langle \mathcal{B}_2, \sim b \rangle$ has two defeaters that can be introduced:
$\langle \mathcal{C}_1, \sim f \rangle = \langle \{(\sim f \prec g, h), (h \prec j)\}, \sim f \rangle$ (proper defeater) and
$\langle \mathcal{C}_2, \sim f \rangle = \langle \{(\sim f \prec i)\}, \sim f \rangle$ (blocking defeater.)
Thus, one of the lines is split in two argumentation lines. The argument $\langle \mathcal{C}_1, \sim f \rangle$ has a blocking defeater that can be introduced in the line: $\langle \mathcal{D}_1, \sim h \rangle = \langle \{(\sim h \prec k)\}, \sim h \rangle$. Finally, observe that both $\langle \mathcal{D}_1, \sim h \rangle$ and $\langle \mathcal{C}_2, \sim f \rangle$ have a blocking defeater, but they cannot be introduced, because they make the argumentation line not acceptable. The dialectical tree for $\langle \mathcal{A}, a \rangle$ is shown in Figure 9. ∎

**Observation** *5.1*
*A subtree of a dialectical tree (i.e., a node with all its descendants) is not always a dialectical tree. Suppose we build an acceptable argumentation line where a defeater $\langle \mathcal{A}, h \rangle$ will not be included because it would make the line unacceptable. There might be a subsequence of the mentioned line where the same defeater could be included, as the following example shows.*

**Example** *5.2*
*Consider the de.l.p. $\mathcal{P}= \{ \ (a \prec b), \ (\sim a \prec c), \ (a \prec f), \ (b), \ (c), \ (d) \ \}$.*
*Here, $\langle \{\sim a \prec c\}, \sim a \rangle$ and $\langle \{a \prec b\}, a \rangle$ are blocking defeaters, and $\langle \{\sim a \prec c\}, \sim a \rangle$ and $\langle \{a \prec f\}, a \rangle$ are also blocking defeaters. Consider the following dialectical tree, with only one argumentation line*

$$\langle \{a \prec b\}, a \rangle$$
$$\uparrow$$
$$\langle \{\sim a \prec c\}, \sim a \rangle$$

*The argument structure $\langle \{a \prec f\}, a \rangle$ cannot be included in the tree as a defeater for $\langle \{\sim a \prec c\}, \sim a \rangle$, because a blocking-blocking situation occurs, and the argument structure $\langle \{a \prec b\}, a \rangle$ cannot be included as a defeater for $\langle \{\sim a \prec c\}, \sim a \rangle$, because it was already used in the argumentation line.*

*The node $\langle \{\sim a \prec c\}, \sim a \rangle$ is a subtree of the one above, but it is not a dialectical tree because $\langle \{a \prec f\}, a \rangle$ and $\langle \{a \prec b\}, a \rangle$, which are defeaters for $\langle \{\sim a \prec c\}, \sim a \rangle$, are not in the tree.* ■

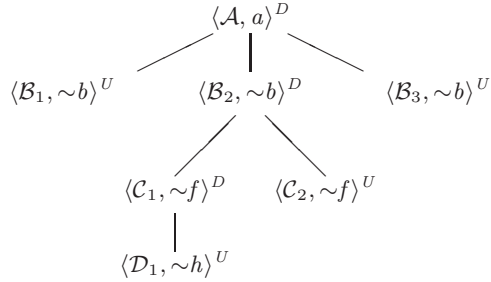

Fig. 10. Marked dialectical tree for Example 5.1

In order to decide whether the root of a dialectical tree is defeated, a marking process will be defined. Nodes will be recursively marked as "$D$" (*defeated*) or "$U$" (*undefeated*) as follows:

**Procedure** *5.1* (*Marking of a dialectical tree*)
Let $\mathcal{T}_{\langle \mathcal{A}, h \rangle}$ be a dialectical tree for $\langle \mathcal{A}, h \rangle$. The corresponding marked dialectical tree, denoted $\mathcal{T}^*_{\langle \mathcal{A}, h \rangle}$, will be obtained marking every node in $\mathcal{T}_{\langle \mathcal{A}, h \rangle}$ as follows:

1. All leaves in $\mathcal{T}_{\langle \mathcal{A}, h \rangle}$ are marked as "$U$"s in $\mathcal{T}^*_{\langle \mathcal{A}, h \rangle}$.
2. Let $\langle \mathcal{B}, q \rangle$ be an inner node of $\mathcal{T}_{\langle \mathcal{A}, h \rangle}$. Then $\langle \mathcal{B}, q \rangle$ will be marked as "$U$" in $\mathcal{T}^*_{\langle \mathcal{A}, h \rangle}$ iff every child of $\langle \mathcal{B}, q \rangle$ is marked as "$D$". The node $\langle \mathcal{B}, q \rangle$ will be marked as "$D$" in $\mathcal{T}^*_{\langle \mathcal{A}, h \rangle}$ iff it has at least a child marked as "$U$".

This procedure suggests a bottom-up marking process, through which we are able to determine the marking of the root of a dialectical tree. Figure 10 shows the dialectical tree of Figure 9 after applying the marking procedure.

The notion of warrant will be defined in terms of a marked dialectical tree as follows.

**Definition** *5.2* (*Warranted literals*)
Let $\langle \mathcal{A}, h \rangle$ be an argument structure and $\mathcal{T}^*_{\langle \mathcal{A}, h \rangle}$ its associated marked dialectical tree. The literal $h$ is warranted iff the root of $\mathcal{T}^*_{\langle \mathcal{A}, h \rangle}$ is marked as "$U$". We will say that $\mathcal{A}$ is a warrant for $h$.

**Proposition** *5.1*
If a literal $q$ has a strict derivation from a *de.l.p.* $\mathcal{P}$, then, $q$ is warranted.

**Proof:** By Observation 3.1, if there exists a strict derivation for $q$ from $\mathcal{P}$, then there exists a unique argument structure for $q$: $\langle \emptyset, q \rangle$. By Proposition 3.1, there exists no possible counter-argument for $\langle \emptyset, q \rangle$. Therefore, there will be no defeaters for $\langle \emptyset, q \rangle$, and then $q$ will be a warranted. ■

In other approaches (Dung, 1995; Toni & Kakas, 1995; Antoniou *et al.*, 2000a; Pollock, 1995; Prakken & Sartor, 1997), the same idea that '*an argument $\mathcal{A}$ will be defeated if there exists at least one defeater for it that it is not defeated*' is used, but in these approaches the notions of argument, defeater, or the argumentation process differ from ours. For example, in (Toni & Kakas, 1995), a similar tree structure was developed for computing the acceptability semantics for negation as failure, however, for them arguments correspond to a set of default negated literals. In (Prakken & Sartor, 1997) a dialogue tree is used, and in (Pollock, 1995) an "Inference Graph" was introduced. See Section 8 for further discussion on the related approaches.

It is interesting to note that the notions of acceptable argumentation line and the dialectical tree provide a flexible structure for defining different *argumentation protocols* when considering different strategies for accepting defeaters during argumentation. This is an advantage over other formalisms where changing the protocol means changing the whole system.

Based on the notion of warrant, we will define a *modal operator of belief* "$B$", where $Bh$ means $h$ is warranted, and $\neg Bh$ means $h$ is not warranted.

**Definition** *5.3* (*Answer to queries*)
The answers of a DeLP interpreter can be defined in terms a modal operator $B$. In terms of $B$, there are four possible answers for a query $h$:

- YES, if $Bh$ ($h$ is warranted)
- NO, if $B\overline{h}$ (the complement of $h$ is warranted)
- UNDECIDED, if $\neg Bh$ and $\neg B{\sim}h$ (neither $h$ nor ${\sim}h$ are warranted.)[3]
- UNKNOWN, if $h$ is not in the language of the program.

---

[3] Observe that the symbol "$\neg$" corresponds to classical negation in the meta-language of the modal operator

**Example** *5.3*
Consider the de.l.p. of Example 5.1. There, the literal "w" is not in the language of the program, so the answer for the query "w" is UNKNOWN. The answer for "a" is UNDECIDED because as shown by the dialectical tree of Figure 10, the literal "a" is not warranted, and the literal "∼a" is also not warranted because there is no argument structure supporting it. The argument $\langle \mathcal{B}_1, \sim b \rangle$ has no defeaters, so "∼b" is warranted (B∼b) and the answer for query "∼b" is YES. Since B∼b, then the answer for query "b" is NO. ∎

**Example** *5.4*
Regarding the de.l.p. $\mathcal{P}_{2.1}$, the answer for $flies(tina)$ is YES, and the answer for $\sim flies(tina)$ is NO.
Considering the program $\mathcal{P}_{2.2}$, the answer for $pacifist(nixon)$ is UNDECIDED, and the answer for $\sim pacifist(nixon)$ is also UNDECIDED.
In the case of example $\mathcal{P}_{2.4}$, the answer for $buy\_stocks(acme)$ is YES, and the answer for $buy\_stocks(alfa)$ is UNKNOWN. ∎

**Example** *5.5*
From the de.l.p. $\mathcal{P}_{2.2}$ of Example 2.2 the following argument structures can be obtained: $\langle \mathcal{A}_1, has\_a\_gun(nixon) \rangle$, $\langle \mathcal{A}_2, \sim has\_a\_gun(nixon) \rangle$, $\langle \mathcal{A}_3, \sim pacifist(nixon) \rangle$, and $\langle \mathcal{A}_4, pacifist(nixon) \rangle$, where

$$\mathcal{A}_1 = \{ has\_a\_gun(nixon) \prec lives\_in\_chicago(nixon) \}$$
$$\mathcal{A}_2 = \left\{ \begin{array}{l} \sim has\_a\_gun(nixon) \prec lives\_in\_chicago(nixon), pacifist(nixon) \\ pacifist(nixon) \prec quaker(nixon) \end{array} \right\}$$
$$\mathcal{A}_3 = \{ \sim pacifist(nixon) \prec republican(nixon) \}$$
$$\mathcal{A}_4 = \{ pacifist(nixon) \prec quaker(nixon) \}$$

Here, $\langle \mathcal{A}_3, \sim pacifist(nixon) \rangle$ is a blocking defeater for $\langle \mathcal{A}_4, pacifist(nixon) \rangle$, and vice versa, therefore in DeLP the answer for the query '$pacifist(nixon)$' will be undecided. The argument structure $\langle \mathcal{A}_2, \sim has\_a\_gun(nixon) \rangle$ is a proper defeater for $\langle \mathcal{A}_1, has\_a\_gun(nixon) \rangle$. Observe that $\langle \mathcal{A}_3, \sim pacifist(nixon) \rangle$ is a blocking defeater for $\langle \mathcal{A}_2, \sim has\_a\_gun(nixon) \rangle$, since $\langle \mathcal{A}_4, pacifist(nixon) \rangle$ is a subargument of $\langle \mathcal{A}_2, \sim has\_a\_gun(nixon) \rangle$.

An argument behaving like $\langle \mathcal{A}_2, \sim has\_a\_gun(nixon) \rangle$ is called in (Makinson & Schlechta, 1991) a "zombie argument": it is not 'alive' because it is blocked by the argument $\langle \mathcal{A}_3, \sim pacifist(nixon) \rangle$, but it is not 'fully dead' because it is defeating the argument $\langle \mathcal{A}_1, has\_a\_gun(nixon) \rangle$. As stated in (Prakken & Vreeswijk, 2000), in such a case, for several argumentation formalisms, neither of the three arguments, $\mathcal{A}_1$, $\mathcal{A}_2$, $\mathcal{A}_3$, is a warrant.

In DeLP $\langle \mathcal{A}_1, has\_a\_gun(nixon) \rangle$ is defeated by $\langle \mathcal{A}_2, \sim has\_a\_gun(nixon) \rangle$, but $\langle \mathcal{A}_2, \sim has\_a\_gun(nixon) \rangle$ is in turn defeated by $\langle \mathcal{A}_3, \sim pacifist(nixon) \rangle$, reinstating $\langle \mathcal{A}_1, has\_a\_gun(nixon) \rangle$. The argument $\langle \mathcal{A}_4, pacifist(nixon) \rangle$ cannot be used to defeat $\langle \mathcal{A}_3, \sim pacifist(nixon) \rangle$ because the argumentation line will not be acceptable (see condition 4 of Definition 4.7). Therefore, $\mathcal{A}_1$ is a warrant for $has\_a\_gun(nixon)$. ∎

**Example** *5.6*

$$\mathcal{P}_{5.6} = \left\{ \begin{array}{ll} \sim p \leftarrow f & p \prec d, h \\ d & \sim p \prec d, h, \sim a \\ h & \sim a \prec e \\ e & a \prec e, f \\ & f \prec d \end{array} \right\}$$

From the de.l.p. above the following argument structures can be obtained:

$$\langle \mathcal{B}, p \rangle = \langle \{p \prec d, h\}, p \rangle$$
$$\langle \mathcal{C}, \sim p \rangle = \langle \{(\sim p \prec d, h, \sim a), (\sim a \prec e)\}, \sim p \rangle$$
$$\langle \mathcal{A}, a \rangle = \langle \{(a \prec e, f), (f \prec d)\}, a \rangle$$

We will first consider a potential warrant for literal $p$. $\langle \mathcal{B}, p \rangle$ has the proper defeater $\langle \mathcal{C}, \sim p \rangle$ defeated in turn by its proper defeater $\langle \mathcal{A}, a \rangle$ in $\sim a$. If the argumentation line $\Lambda_1 = [\langle \mathcal{B}, p \rangle, \langle \mathcal{C}, \sim p \rangle, \langle \mathcal{A}, a \rangle]$ were acceptable, literal "$p$" would be warranted.

Next, we will consider a potential warrant for literal $\sim p$. As stated above, $\langle \mathcal{C}, \sim p \rangle$ has the proper defeater $\langle \mathcal{A}, a \rangle$, but note that $\langle \mathcal{A}, a \rangle$ is also defeated by $\langle \mathcal{B}, p \rangle$ (using the strict rule "$\sim p \leftarrow f$"). Consider $\Lambda_2 = [\langle \mathcal{C}, \sim p \rangle, \langle \mathcal{A}, a \rangle, \langle \mathcal{B}, p \rangle]$. Although $\langle \mathcal{C}, \sim p \rangle$ defeats $\langle \mathcal{B}, p \rangle$, it cannot be introduced in $\Lambda_2$ because of the circularity restriction. If $\Lambda_2$ were to be acceptable, the literal $\sim p$ would now be warranted.

Therefore, accepting $\Lambda_1$ for $p$ and $\Lambda_2$ for $\sim p$ would render both literals warranted. This will no happen in DeLP because neither $\Lambda_1$ nor $\Lambda_2$ satisfy the concordance restriction and therefore they are not acceptable argumentation lines. Observe that $\langle \mathcal{B}, p \rangle$ and $\langle \mathcal{A}, a \rangle$ are not concordant, and $\langle \mathcal{B}, p \rangle$ and $\langle \mathcal{C}, \sim p \rangle$ are not concordant either. ∎

### 5.1 The Warrant Procedure with pruning

In order to decide whether a literal $h$ is warranted from a de.l.p. $\mathcal{P}$, the warrant procedure has to find an argument structure $\langle \mathcal{A}, h \rangle$ and, as established by Definition 5.2, the root of $\mathcal{T}^*_{\langle \mathcal{A}, h \rangle}$ has to be marked as "$U$". We will introduce in this section a procedure for deciding whether a given literal is warranted. This procedure will not explore, in general, the whole dialectical tree, and answers will therefore be computed in a more efficient way.

Given a program $\mathcal{P}$, there could be several argument structures $\langle \mathcal{A}_1, h \rangle, \ldots, \langle \mathcal{A}_i, h \rangle$ for a literal $h$. However, the warrant procedure will not construct all the possible argument structures for $h$; it will consider each one of them in turn, exploring the associated dialectical tree. This optimization is similar in spirit to the one found in OSCAR (Pollock, 1996).

Observe that a marked dialectical tree $\mathcal{T}^*_{\langle \mathcal{A}, a \rangle}$, like the one in Figure 11 (left), resembles the *minimax* tree used in Artificial Intelligence for game trees. Here, instead of nodes marked with 1 or -1, the tree has nodes marked "$D$", or "$U$".

Note also that during the marking of the dialectical tree, some nodes are not contributing to the decision procedure (the marking), i.e. are such that they could be either "$U$" or "$D$" without changing the marking of the dialectical tree's root. For example, in Figure 11 (left) the left-most child of the root is a "$U$", so the root is a "D", no matter what the marking of the other two children of the root are. Hence, such a *don't-care node* obviously belongs to a branch that may be pruned. This pruning process is similar to the α-β pruning of a mini-max tree.

Clearly, during the marking procedure, once a node is labeled "U" all of its siblings can be pruned. Figure 11 (left) shows a marked dialectical tree for argument structure $\langle \mathcal{A}, a \rangle$ of Example 5.1 and the pruned tree in depth-first order (right.)

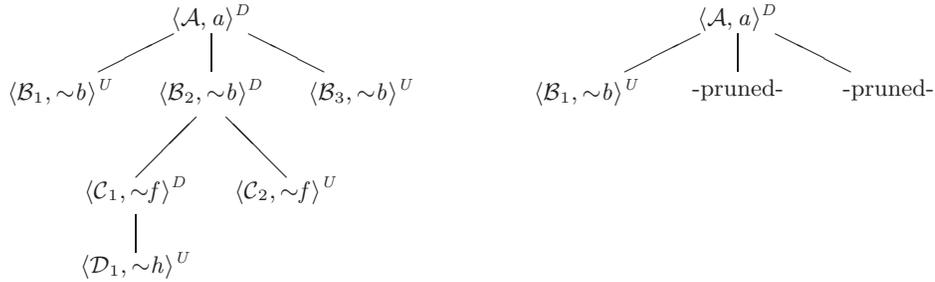

Fig. 11. Marked Dialectical tree for example 5.1 (left) and pruned (right)

Given a query $q$ the warrant procedure first will try to generate an argument structure $\mathcal{A}_1$ for $q$. If $\mathcal{A}_1$ for $q$ is found, then the warrant procedure will try to build a defeater $\mathcal{A}_2$ for some counter-argument point in $\mathcal{A}_1$ (see the example below.) If such defeater exists, it will try to build a defeater $\mathcal{A}_3$ for $\mathcal{A}_2$, and so on, building in this form an argumentation line. Thus, a dialectical tree will be generated in depth-first manner, considering (from left to right) every acceptable argumentation line.

In a dialectical tree there are as many argumentation lines as leaves in the tree, and each of them could finish in a supporting or an interfering argument. Example 5.7 shows how a dialectical tree is constructed in a depth-first manner, considering supporting and interfering arguments for each possible argumentation line, and how the marking procedure and pruning is done while building the tree.

**Example** *5.7*
*Suppose that, in order to find a warrant for $h_1$, the argument $\mathcal{A}_1$ is found, and the acceptable argumentation line [ $\langle \mathcal{A}_1, h_1 \rangle$, $\langle \mathcal{A}_2, h_2 \rangle$, $\langle \mathcal{A}_3, h_3 \rangle$, $\langle \mathcal{A}_4, h_4 \rangle$, $\langle \mathcal{A}_5, h_5 \rangle$] is built, see Figure 12 (i.) In this situation, the acceptable argumentation line ends with the supporting argument $\mathcal{A}_5$, so the marking procedure establishes that $\langle \mathcal{A}_1, h_1 \rangle$ is –up to this point– a "U". However, the warrant process cannot finish there because there could be more defeaters to consider. Therefore, the process will continue expanding other argumentation lines.*

*First, note that although there could be more defeaters for $\mathcal{A}_4$, considering them will not change $\mathcal{A}_4$'s status because of $\mathcal{A}_5$. Therefore, the tree can be pruned at that point without considering further defeaters for $\mathcal{A}_4$.*

*However, the previous analysis does not apply to $\mathcal{A}_3$, because if an undefeated defeater is found for it, the mark of $\mathcal{A}_3$ could change. It is for this reason that* the procedure will look for any other possible defeater $\mathcal{A}_4{'}$ for $\mathcal{A}_3$, creating a new argumentation line, see Figure 12-ii.

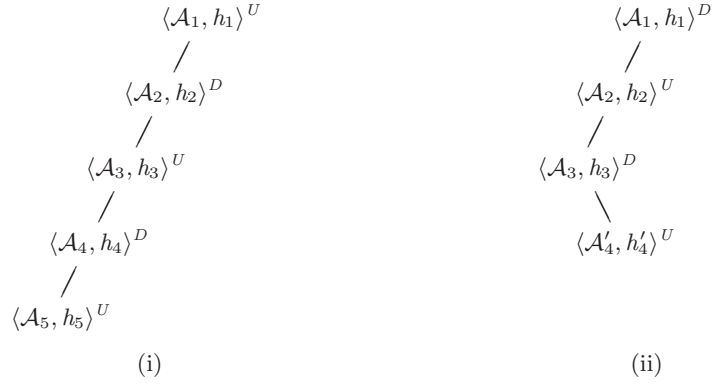

Fig. 12. Argumentation lines of Example 5.7

If a defeater $\mathcal{A}_4{'}$ is found (with no defeaters), then the argumentation line will end with an interfering argument, and therefore $\mathcal{A}_1$ will be a "D", see Figure 12-ii. Again, pruning could be effected, because although there could be more defeaters for $\mathcal{A}_3$, they cannot modify its status. However, there might be another defeater $\mathcal{A}_3{'}$ for $\mathcal{A}_2$, creating, in that case, a new argumentation line.    ∎

Figure 13 shows a PROLOG-like specification of the top level of the warrant procedure with pruning. Predicates `warrant/2` and `defeated/2` specify how to perform the dialectical analysis. That is, a query `Q` will be warranted if an argument `A` for `Q` is found, and `A` is not defeated. The predicate `find_argument/2` (not developed in the figure) simply builds an argument for a given query.

The predicate `defeated/2`, receives an argument `A` and an argumentation line `ArgLine`, and tries to find a defeater `D` for `A`, checking that `D` is acceptable as part of the argumentation line. If `acceptable/3` succeeds, then it returns `NewLine` adding `D` to `ArgLine`. Since the argument `A` will be defeated if there exists a defeater that is in turn not defeated, then finally a call to `\+ defeated(D,NewLine)` is made.

The predicate `find_defeater/2` calls `find_counterarg/2` that looks for an argument `C` that counter-argues `A` with a disagreement sub-argument `SubA`. The argument `C` will be a defeater for `A` if `SubA` is not better than `C`. The predicate `better/2` succeeds when the first argument is better than the second regarding the chosen comparison criterion. Observe finally that the pruning is performed calling `defeated/1` recursively with Prolog's negation as failure "`\+`".

```
warrant(Q,A):-                          % Q is a warranted literal
  find_argument(Q,A),                   % if A is an argument for Q
  \+ defeated(A,[support(A,Q)]).        % and A is not defeated

defeated(A,ArgLine):-                   % A is defeated
  find_defeater(A,D,ArgLine),           % if there is a defeater D for A
  acceptable(D,ArgLine,NewLine),        % acceptable within the line
  \+ defeated(D,NewLine).               % and D is not defeated

find_defeater(A,C):-                    % C is a defeater for A
  find_counterarg(A,C,SubA),            % if C counterargues A in SubA
  \+ better(SubA,C).                    % and SubA is not better than C
```

Fig. 13. Specification of the Warrant Procedure with Pruning

## 6 DeLP Extensions

### *6.1 DeLP with Default Negation*

As discussed in (Alferes *et al.*, 1996), logic programs, deductive databases, and more generally non-monotonic theories, use various forms of *default negation*, "*not F*", whose major distinctive feature is that '*not F*' is assumed by default, i.e., it is assumed in the absence of sufficient evidence to the contrary. In DeLP "absence of sufficient evidence" means "there is not warrant". Therefore, the default negation '*not F*' will be assumed when the literal $F$ is not warranted.

We will discuss here briefly how to extend DeLP for using default negation. A more detailed paper with the definition of extended DeLP, and a comparison with other approaches is in preparation.

When DeLP is extended to consider default negation, some characteristics of the formalism just described are affected. For a correct treatment of default negation in DeLP, further considerations will be required.

Default negation will be allowed only preceding literals in the body of defeasible rules, e.g., '$\sim cross\_railway\_tracks \prec not \sim train\_is\_coming$[4]', and defeasible rules that use default negation will be called *extended defeasible rules*. The reason not allowing default negation in strict rules is twofold. On one hand, a strict rule '$p \leftarrow not\ q$' is not completely strict, because the head '$p$' will be derived assuming '$not\ q$'. On the other hand, the set $\Pi$ of strict rules and facts could become a contradictory set in many cases. An Extended Defeasible Logic Program will be then, a set of Facts, Strict Rules and Extended Defeasible Rules.

Since the decision of assuming an extended literal "*not L*" will be carried out by the dialectical process, the definition of defeasible derivation (Definition 2.5) is modified acordingly in *extended* DeLP. The change reflects that when an extended literal is found in the body of a rule, that literal will be ignored:

---

[4] Adapted from an example attributed to John McCarthy in (Gelfond & Lifschitz, 1990).

**Definition** *6.1* (*Extended Defeasible Derivation*)
Let $\mathcal{P}=(\Pi, \Delta)$ be an extended defeasible logic program and $L$ a ground literal. A defeasible derivation of $L$ from $\mathcal{P}$, denoted $\mathcal{P} \mid\!\sim L$, consists of a finite sequence $L_1, L_2, \ldots, L_n = L$ of ground literals, and each literal $L_i$ is in the sequence because:

(a) $L_i$ is a fact in $\Pi$, or
(b) there exists a rule $R_i$ in $\mathcal{P}$ (strict or defeasible) with head $L_i$ and body $B_1, B_2, \ldots, B_k$ and every literal of the body, except the ones preceded by default negation, is an element $L_j$ of the sequence appearing before $L_i$ ($j < i$.)

The definition of argument structure is also extended in order to avoid the introduction of self-defeating arguments, shown in the following example. Consider the set of defeasible rules: $\mathcal{A}=\{(a \prec b), (b \prec \text{not } a)\}$. From any *de.l.p.* including those rules, it is possible to obtain a defeasible derivation for "$a$", assuming "*not a*". However, an argument structure like $\langle \mathcal{A}, a \rangle$ would be a new kind of self-defeating argument that we would like to avoid. Observe the new condition "2" below.

**Definition** *6.2* (*Extended Argument Structure*)
Let $h$ be a literal, and $\mathcal{P}=(\Pi, \Delta)$ an extended defeasible logic program. An *argument structure* $\langle \mathcal{A}, h \rangle$ for a ground literal $h$, is a set of extended defeasible rules of $\Delta$, such that:

1. there exists a defeasible derivation for $h$ from $\Pi \cup \mathcal{A}$.
2. if $L$ is a literal in the defeasible derivation of $h$, then there is no defeasible rule in $\mathcal{A}$ containing "*not L*" in its body.
3. $\Pi \cup \mathcal{A}$ is non-contradictory, and
4. $\mathcal{A}$ is minimal: there is no proper subset $\mathcal{A}'$ of $\mathcal{A}$ such that $\mathcal{A}'$ satisfies conditions (1) and (3).

In extended DeLP, default negated literals will be another point of attack in an argument. Phan Dung in (Dung, 1993a) points out that "*default negated literals are assumptions on which the derivation is based. The acceptance of the derivation depends on the acceptance of these assumptions*". In his work, Dung defines a notion of *ground attack*: an argument $\mathcal{A}'$ for $l$, is a ground attack for $\mathcal{A}$, if $\mathcal{A}$ contains a default negated literal *not l*. That is, an argument based on an assumption "*not l*" could be attacked by an argument that supports the literal "*l*". We will extend our notion of defeat, incorporating Dung's notion of ground attack. Something similar was done in (Prakken & Sartor, 1997).

We say that *an argument structure $\langle \mathcal{A}, h \rangle$ is an attack to an assumption of $\langle \mathcal{B}, q \rangle$*, if the extended literal "*not h*" is in the body of a defeasible rule in $\mathcal{B}$. The notion of defeater will be extended considering this new kind of attack.

**Definition** *6.3*
An argument structure $\langle \mathcal{A}_1, h_1 \rangle$ is a defeater for $\langle \mathcal{A}_2, h_2 \rangle$, if and only if either:
(*a*) $\langle \mathcal{A}_1, h_1 \rangle$ is a proper defeater for $\langle \mathcal{A}_2, h_2 \rangle$, or
(*b*) $\langle \mathcal{A}_1, h_1 \rangle$ is a blocking defeater for $\langle \mathcal{A}_2, h_2 \rangle$, or
(*c*) $\langle \mathcal{A}_1, h_1 \rangle$ is an attack to an assumption of $\langle \mathcal{A}_2, h_2 \rangle$.

With this new definition of defeater, default negated literals become new points of attack. Thus, when the dialectical analysis is carried out, default negated literals could be defeated by arguments. It's easy to see that DeLP negation satisfies the *coherence principle* established in (Alferes & Pereira, 1994; Pereira & Alferes, 1994): If "$\sim p$" is warranted, then "*not p*" can be assumed.

We claim that both negations are needed for representing knowledge in a natural manner. However, some approaches in the literature (Kakas *et al.*, 1994; Dimopoulos & Kakas, 1995; Xianchang Wang, 1997) have tried to define default negation in terms of strong negation. Here follows some proposed transformations and counterexamples showing why they fail.

In the approach of Logic Programming without Default Negation (Kakas *et al.*, 1994; Dimopoulos & Kakas, 1995), a priority relation between rules is used for deciding between competing rules. In their approach they remove default negation using the following transformation: the rule "$r_0 : p \leftarrow q, not\ s$" is transformed into two rules, "$r_1 : p \leftarrow q$" and "$r_2 : \sim p \leftarrow s$", with $r_1 < r_2$. Hence, when $s$ is not derivable, the rule $r_2$ cannot be used, and there is a derivation for $p$. On the other hand when $s$ is derivable, rule $r_2$ blocks $r_1$.

However, in this approach, when $s$ becomes derivable after the transformation, the literal $\sim p$, that was not derivable from the original program is now derivable in the transformed one. This new derivable literal may cause unexpected results, as shown in Example 6.1 where we compare a *de.l.p.* $\mathcal{P}$ with default negation, and the program $\mathcal{P}'$ obtained with the transformation cited above. The program $\mathcal{P}'$ has the priority: $(\sim p \prec r) > (p \prec q)$.

**Example** *6.1*

| $\mathcal{P}$ | $\mathcal{P}'$ |
|---|---|
| $p \prec q, not\ s$ | $p \prec q$ |
|  | $\sim p \prec s$ |
| $q$ | $q$ |
| $s$ | $s$ |
| $a \prec q$ | $a \prec q$ |
| $\sim a \prec \sim p$ | $\sim a \prec \sim p$ |

∎

Observe that from $\mathcal{P}$ there is no argument for $p$, there is no argument for $\sim p$, and the literal $a$ is warranted. However, in the transformed program $\mathcal{P}'$ the literal $\sim p$ is warranted and there is no warrant for the literal $a$, because $\sim p$ allows to build an argument for $\sim a$ that defeats the argument for $a$. Further comments on the transformation cited above were reported in (Xianchang Wang, 1997).

In (Xianchang Wang, 1997), where Priority Logic Programming was defined, another transformation is given: a rule "$p \leftarrow not\ q$" is transformed to "$p \leftarrow \overline{q}$", where "$\overline{q}$" is a new symbol. In addition, for every literal $p$ of the program, two new rules are generated: "$r_1 : p \leftarrow p$" and "$r_2 : \overline{p}$", with $r_2 < r_1$. We refer the interested reader to (Xianchang Wang, 1997) for the details of the transformation. Here follows an example of a *de.l.p.* $\mathcal{P}$ and its transformation $\mathcal{P}'$.

**Example** *6.2*

| $\mathcal{P}$ | $\mathcal{P}'$ |
|---|---|
| $p \leftarrow \mathit{not}\, q$ | $p \leftarrow \overline{q}$ |
| | $\overline{q}$ |
| | $\overline{p}$ |
| | $p \leftarrow p$ |
| | $q \leftarrow q$ |

∎

In Example 6.2, from the transformed program $\mathcal{P}'$ new literals will be derived that may interact with other parts of the program.

### 6.2 DeLP with presumptions

A defeasible rule with an empty body is called a *presumption* (Nute, 1988). In our approach a rule like "$a \prec$" would express that "*there are (defeasible) reasons to believe in a*".

Extending DeLP to consider presumptions is straightforward. We will show that only slight modifications need to be made in the formalism. An extended defeasible logic program will be a set of facts, strict rules, defeasible rules and presumptions. We will denote with $\Delta^+$ the set of defeasible rules, and presumptions. The definition of defeasible derivation is the only one that has to be extended in order to consider presumptions. In Definition 2.5, condition (a) has to be changed to: "$L_i$ is a fact or a presumption".

One major difference with respect to a regular *de.l.p.* is that presumptions are defeasible rules without body. Given an extended *de.l.p.* $(\Pi,\Delta^+)$ with no facts, defeasible derivations and arguments can still be obtained. For example, from the *de.l.p.* $\mathcal{P}_1 = \{(b \prec ), (a \prec b)\}$ a defeasible derivation for "$a$" can be obtained. Thus, argument structures could be based on facts, on presumptions, or both.

Since presumptions are a special case of defeasible rules, the notion of argument structure remains intact. However, given an argument structure $\langle \mathcal{A}, h \rangle$, the set $\mathcal{A}$ will be a set of defeasible rules that could include presumptions ($\mathcal{A} \subseteq \Delta^+$). The definitions of disagreement, counter-argument, defeater, dialectical tree, and the warrant procedure are not affected by the inclusion of presumptions.

The comparison criterion could be affected. As the following example shows, the specificity criterion defined in this paper has some problems when the argument contains presumptions.

**Example** *6.3*
Consider the extended *de.l.p.* $(\Pi,\Delta^+)$, where $\Pi= \{f\}$, and $\Delta^+ = \{(a \prec p,f), (p \prec ), (\sim a \prec f), (a \prec t), (t \prec )\}$. The following argument structures can be obtained:

$$\langle \mathcal{A}_1, a \rangle = \langle \{(a \prec t), (t \prec )\}, a \rangle$$
$$\langle \mathcal{A}_2, \sim a \rangle = \langle \{\sim a \prec f\}, \sim a \rangle$$
$$\langle \mathcal{A}_3, a \rangle = \langle \{(a \prec p,f), (p \prec )\}, a \rangle$$

Observe that $\langle \mathcal{A}_2, \sim a \rangle$ is based on the fact $f$, and $\langle \mathcal{A}_1, a \rangle$ is based on the presumption $t$. Clearly, an argument based on facts should be preferable to one based on presumptions. In this case Definition 3.5 behaves as expected, because it states that $\langle \mathcal{A}_2, \sim a \rangle$ is more specific than $\langle \mathcal{A}_1, a \rangle$.

However, in other cases, this definition does not behave correctly. Observe now that $\langle \mathcal{A}_2, \sim a \rangle$ is based on the fact $f$, $\langle \mathcal{A}_3, a \rangle$ is based on the fact $f$ and the presumption $p$, i. e. $\langle \mathcal{A}_3, a \rangle$ is using more information. Here, Definition 3.5 states that $\langle \mathcal{A}_2, \sim a \rangle$ and $\langle \mathcal{A}_1, a \rangle$ are incomparable. The reason for this is that presumptions do not have a body and therefore the set $H = \{f\}$ activates $\mathcal{A}_2$. Other examples were analyzed in (García, 2000).

If the comparison criterion used is based on rules priorities, then the criterion has to find the way of preferring a fact over a presumption. Otherwise, an argument based on a fact and an argument based on a presumption (like $\langle \mathcal{A}_2, \sim a \rangle$ and $\langle \mathcal{A}_1, a \rangle$ in the example above) will be of equal strength.

One simple way of solving the problems mentioned above is establishing that arguments based on facts will be preferable to arguments based on presumptions. The extension of the comparison criteria to consider presumptions is currently under study.

## 7 Implementation and Applications

An interpreter of DeLP was implemented in PROLOG, and can be used through the web (see http://cs.uns.edu.ar/∼ ajg/DeLP.html). Also an abstract machine called JAM (Justification Abstract Machine) (García, 1997) has been designed for the implementation of DeLP, as an extension of the Warren's abstract machine (WAM). A prototype implementation of the JAM as a virtual machine was also developed, and is subject of future research.

Applications that deal with incomplete and contradictory information can be easily modeled using DeLP programs. The defeasible argumentation basis of DeLP allows the building of applications for dynamic domains, where information may change. Thus, Defeasible Logic Programming can be used for representing knowledge and for providing an inference engine in many applications. A concrete application of DeLP was (García *et al.*, 2000), where a multi-agent system for the stock market domain was developed. The application consists of several deliberative agents for monitoring the stock market and performing actions based on the retrieved information. The agents reason using DeLP, and are capable of formulating arguments and counterarguments in order to decide whether to buy or sell some stock. Other applications are in progress.

## 8 Related Work

DeLP combines Defeasible Argumentation and Logic Programming. In both areas there have been developed several related approaches. We will comment first the differences with other Defeasible Reasoning formalisms and then with extentions of Logic Programming that are related with our work.

### *8.1 Defeasible Logic and Argumentation*

Nute's d-Prolog (Nute, 1988; Nute, 1994) was the first to introduce defeasible reasoning programming with specificity. d-Prolog syntax has strict and defeasible rules and strong negation. However, d-Prolog's rules do not allow default negation. The language of d-Prolog provides facilities to define *defeater rules* like *"sick birds do not fly"*. The purpose of defeater rules is to account for the exceptions to defeasible rules. However, in (Antoniou *et al.*, 2001) it is shown that defeaters can be simulated by means of strict and defeasible rules (in Nute's sense).

DeLP does not need to be supplied with *defeater rules*. The system will find the counterarguments among the arguments it is able to build, and will decide on the defeat relation using a comparison criterion. Thus, in DeLP the programmer does not need to encode explicit exceptions.

One important difference between d-Prolog and our approach is the way in which contradictory conclusions are treated. In d-Prolog there is no notion of argument. In order to decide between two contradictory conclusions, d-Prolog compares only one pair of rules, whereas in DeLP the two arguments that support those conclusions are compared. Comparing only a pair of rules may be problematic as we show next.

Consider the program $\mathcal{P}_1 = \{(a \prec b), b, (c \prec d), d, (h \leftarrow a), (\sim h \leftarrow c)\}$ of Example 2.3. In d-Prolog the literal $a$ is accepted as proved from $\mathcal{P}_1$ because there is no rule with "$\sim a$" in its head, so no rule that contradicts "$a \prec b$" is found. However, literals $a$ and $c$ disagree: literals $h$ and $\sim h$ are derivable from $\{a, c\} \cup \{(h \leftarrow a), (\sim h \leftarrow c)\}$. In DeLP, the argument $\mathcal{B} = \{(c \prec d)\}$ is a blocking defeater for $\mathcal{A} = \{a \prec b\}$, because $a$ and $c$ disagree. Therefore, $a$ fails to be warranted, and the answer for $a$ is UNDECIDED. Answer for $c$ is also UNDECIDED.

Another problematic situation of comparing two rules without considering the rest of the program follows. Some approaches consider that $R_1 =$"$\sim p \prec q, r$" is better than $R_2=$"$p \prec q$" because the body of $R_1$ has more information. However, it may not be true, depending on the basis for the literal $r$. For instance, consider the program $\mathcal{P}_2 = \{(p \prec q), q, (\sim p \prec q, r), (r \leftarrow q)\}$. Here, $r$ is obtained strictly from $q$, so it's not true that $R_1$ is based on more information than $R_2$. Both rules have the same basis: the literal $q$.

A major difference is that in d-Prolog there is no dialectical analysis, and no treatment for circular argumentation lines. The interested reader is referred to (Prakken & Vreeswijk, 2000), where other features of Nute's work are discussed.

Besides Nute's work on Defeasible Logic, recent work by Grigoris Antoniou, David Billington, Michel Maher and Guido Governatori, has extended Nute's approach, see (Antoniou *et al.*, 2000a; Antoniou *et al.*, 1998). Unfortunately, the same problems mentioned above for d-Prolog are inherited there.

The defeasible argumentation formalism developed in (Simari & Loui, 1992) and used here, was inspired in part by Pollock's work in Defeasible Reasoning (Pollock, 1987). However, Pollock has changed the way in which an argument is warranted, adopting a *multiple status assignment approach*[5] (Pollock, 1995; Pollock, 1996). Pollock has developed a computer program in Lisp, called OSCAR (Pollock, 1995) that performs defeasible reasoning. In OSCAR, arguments are sequences of linked *reasons*, and probabilities are used for comparing competing arguments. In a way similar to Nute's defeater rules, explicit 'undercutting' defeaters can be expressed in his language. An inference graph is used by OSCAR for evaluating the status of arguments. Pollock argues that human reasoning is defeasible in two different senses. He distinguishes between 'synchronically defeasible' (a conclusion may be unwarranted relative to a larger set of inputs) and 'diachronically defeasible' (a conclusion may be retracted as a result of further reasoning, without any new input). Hence, in OSCAR an argument may be 'justified' in one stage of reasoning, and unjustified later, without any additional input. However an argument is 'warranted' when the reasoner reaches a stage, where for any new stages of reasoning the argument remains undefeated. This notion of warrant coincides with ours. However, in OSCAR a bottom-up procedure is used for computing justified and warranted arguments.

In (Dung, 1995), P. Dung has proposed a very abstract and general argument-based framework, where he completely abstracts from the notions of argument and defeat. In contrast with our approach of defining an object language for representing knowledge and a concrete notion of argument and defeat, Dung's approach assumes the existence of a set of arguments ordered by a binary relation of defeat. However, he defines various notions of 'argument extensions', which aim to capture various types of defeasible consequence.

Inspired by legal reasoning, H. Prakken and G. Sartor (Prakken & Sartor, 1997) have developed an argumentation system that, like ours, uses the language of extended logic programming. They introduce a *dialectical proof theory* for an argumentation framework fitting the abstract format developed by Dung, Kowalski *et al.* (Dung, 1995; Bondarenko *et al.*, 1993). However, since they are inspired by legal reasoning, the protocol for dispute is rather different from our dialectical tree. A proof of a formula takes the form of a *dialogue tree*, where each branch of the tree is a dialogue between a *proponent* and an *opponent*. Proponent and opponent have different rules for introducing arguments, leading to an asymmetric dialogue. Later, Prakken (Prakken, 1997) generalized the system to default logic's language.

R. Kowalski and F. Toni (Kowalski & Toni, 1996) have outlined a formal theory of argumentation, in which defeasibility is stated in terms of non-provability claims. They argue that defeasible reasoning with rules of the form $P$ <u>if</u> $Q$ can be understood as "exact" reasoning with rules of the form $P$ <u>if</u> $Q$ <u>and</u> $S$ <u>cannot be shown</u>, where $S$ stands for one or more defeasible "non-provability claims". In (Brewka, 2001a), a proposal of a new formal notion of argument systems is given, that focuses on capturing the most revelant aspects of realistic argumentation processes. His main interest is to capture the logic and procedural aspects of argumentation. The underlying language of his approach is preferential default logic.

Other related approaches of defeasible argumentation are by Verheij (Verheij, 1996),

---

[5] Unique and multiple status assignments for arguments are analyzed in depth in (Prakken & Vreeswijk, 2000)

Vreeswijk (Vreeswijk, 1997), Bondarenko (Bondarenko *et al.*, 1997), and Loui (Loui, 1997b). Details of them can be found in the following surveys of defeasible argumentation: (Prakken & Vreeswijk, 2000), and (Chesñevar *et al.*, 2000).

### 8.2 Logic Programming

In (Gelfond & Lifschitz, 1990), *Logic Programming with Classical Negation* was introduced. There, when two complementary literals can be derived, the program becomes "contradictory" and every literal of the program can be derived. Since common sense reasoning is typically based on tentative information, and the representation of this kind of information leads in most cases to inconsistent knowledge bases, an extended logic program usually will derive all of the language. This problem was attacked in (Inoue, 1991), where *Extended Logic Programming with Default Assumptions* is considered. This approach resembles a defeasible argumentation system, but unfortunately no preference criterion for deciding between contradictory explanations was considered.

In (Kakas *et al.*, 1994), a semantics for default negation called "acceptability semantics" was introduced, based on previous works on default negation and abductive logic programming (Eshghi & Kowalski, 1989; Dung, 1991; Kakas *et al.*, 1993). Sets of default negated literals are considered as extensions of the program, and a notion of "attack" between these sets is defined. An extension $H$ is acceptable iff any attack $A$ against $H$ is not acceptable. A fixpoint operator for acceptability is given. They introduce a general theory of acceptability based on a binary relation "attack" that for LP is defined using a priority relation over program rules.

Toni and Kakas in (Toni & Kakas, 1995), developed a proof procedure for the acceptability semantics mentioned above. In their approach a tree structure similar to our dialectical tree was developed. Both approaches share the idea of having a tree where children nodes attack the father node. However, as the intended application of these trees is computing a semantics for default negation, the nodes of their tree are sets of 'abducibles' (default negated literals), whereas a dialectical tree is a tree of arguments. Since they do not consider strong negation, both approaches are difficult to compare.

In (Kakas *et al.*, 1994) and later in (Dimopoulos & Kakas, 1995), "Logic Programming without Negation as Failure" (LPwNF) was introduced. A LPwNF program consists of a set of basic rules $L_0 \leftarrow L_1, \ldots, L_n$ (where $L_i$ are literals that could use strong negation) and a given irreflexive and antisymmetric priority relation among program rules. They claim that default negation can be removed using a program transformation. The problem with this transformation (see Example 6.1) is that new literals are derivable that may cause other derivations to be blocked. Other problems with the transformation were reported in (Xianchang Wang, 1997).

The proof procedure of LPwNF is very similar to the one of d-Prolog. Although in (Dimopoulos & Kakas, 1995) there is no comparison with Defeasible Logic, in (Antoniou *et al.*, 2000a) a comparison among LPwNF, Defeasible Logic, and 'Courteous Logic Programs' is given. The main result of (Antoniou *et al.*, 2000a) is that Defeasible Logic can prove everything that sceptical LPwNF can. In (Gelfond & Son, 1997) a system to 'investigate the methodology of reasoning with prioritized defaults in the language of logic programs under the answer set semantics' was developed. Their system allows the representation of defeasible and strict rules, and the representation of an order among those rules. The way in which defeasible inferences are obtained is very similar to Antoniou *et al.* approach, although no comparison of these two systems is given.

Sometimes, defeasible rules are considered as defaults in a default theory. However, defaults are not defeasible rules, as explained in (Nute, 1994). We will now introduce a more illustrative example adapted from (Covington *et al.*, 1997).

**Example** *8.1*
*Consider the following de.l.p. (left) and the same knowledge represented in a default theory (right).*

$$has\_shell(X) \prec mollusc(X)$$
$$\sim has\_shell(X) \prec cephalopod(X)$$
$$mollusc(X) \leftarrow cephalopod(X)$$
$$cephalopod(fred)$$

$$\frac{mollusc(X):has\_shell(X)}{has\_shell(X)}$$
$$\frac{cephalopod(X):\sim has\_shell(X)}{\sim has\_shell(X)}$$
$$mollusc(X) \leftarrow cephalopod(X)$$
$$cephalopod(fred)$$

*From the DeLP program above, there is an argument for $\sim has\_shell(fred)$ that is more specific than the argument for $has\_shell(fred)$. Hence, there is a warrant for $\sim has\_shell(fred)$. However, in Default Logic there are two extensions: one with $\sim has\_shell(fred)$ and the other with $has\_shell(fred)$. The reason is because the defaults express no connection between cephalopod and mollusc. To capture this connection, the first default should be changed to $\frac{mollusc(X):has\_shell(X) \land \sim cephalopod(X)}{has\_shell(X)}$. That is, the exception must be explicitly encoded in the default. In DeLP, however, the exceptions are discovered by the warrant procedure.* ■

The above comparison seems to be unfair because default logic has no selection mechanism. However, in (Brewka & Eiter, 2000) default logic was extended in order to handle priorities, developing a *Prioritized Default Logic* (PDL). This approach has many properties which are relevant for argumentation, such as explicit representation of preferences and reasoning about these preferences. Although this approach is not explicitly argument-based, prioritized default theories extend default theories adding a strict partial order on defaults, using this ordering to define preferred extensions. PDL satisfies two reasonable principles for preference handling, which distinguishes PDL from other approaches. However, since an ordering of defaults is enforced, problems similar to those mentioned for comparing two rules are also present. Another important difference is that they only consider sets of default rules, without introducing strict rules, as was done here.

In (Brewka, 2001b) a comparison between a variant of Defeasible Logic, called *Ambiguity Propagating* (Antoniou *et al.*, 2000b), and the prioritized version of Well-Founded Semantics for extended logic programs (Brewka, 1996) is given. The paper shows that under the condition that preferences are admitted between defeasible rules only, then all defeasibly provable literals by the defeasible logic variant are true in prioritized well-founded semantics. It also shows that there are some desirable conclusions obtained by well-founded semantics that the variant of defeasible logic cannot obtain.

In Prioritized Logic Program (PLP), a program is a pair $(P, >)$ where $P$ is a finite set of rules of the form "$c \leftarrow a_1, \ldots, a_n, not\ b_1, \ldots, not\ b_m$" and $>$ is an acyclic preference relation on $P$. In PLP, a rule $r$ is said to be defeated by a literal $l$ if $l = b_i$, for some $i \in 1, \ldots, m$. Clearly, if no default negated literals are used in a program, then no rule is defeated. This represents a difference with DeLP because counter-arguments and defeaters are defined in term of strong negation. In PLP, default negated literals are the only point of attack, whereas in DeLP arguments are attacked by other arguments.

The comparison in (Brewka, 2001b) is based on the translation of each defeasible rule "$\{a_1, \ldots, a_n\} \Rightarrow b$" of default logic, to an extended rule with default negation: "$b \leftarrow not{\sim}b, a_1, \ldots, a_n$". However, note that this translation captures only an attack to a rule, and not an attack to an argument (see the example of program $\mathcal{P}_1$ above).

One distinguishing feature of DeLP is the property of argument reinstatement. For example, consider the following *de.l.p.* $\mathcal{P}$:

$$a \prec b$$
$${\sim}a \prec b, c$$
$$c \prec i$$
$${\sim}c \prec i, j$$
$$b$$
$$i$$
$$j$$

Here, the argument $\langle \{a \prec b\}, a\rangle$ for the literal $a$ has a proper defeater, the argument for ${\sim}a$: $\langle \{({\sim}a \prec b, c), (c \prec i)\}, {\sim}a\rangle$. This one is in turn defeated by the proper defeater: $\langle \{{\sim}c \prec i, j\}, {\sim}c\rangle$, that attacks the argument for ${\sim}a$ in the inner point $c$. This third argument reinstates the first, and thus, there is a warrant for $a$. Therefore the set of warranted literals from $\mathcal{P}$ is $W = \{a, {\sim}c, b, i, j\}$

In order to encode the program $\mathcal{P}$ in PLP, we have used the translation suggested in (Brewka, 2001b):

$$r_1: a \leftarrow not\ {\sim}a, b$$
$$r_2: {\sim}a \leftarrow not\ a, b, c$$
$$r_3: c \leftarrow not\ {\sim}c, i$$
$$r_4: {\sim}c \leftarrow not\ c, i, j$$
$$r_5: b$$
$$r_6: i$$
$$r_7: j$$

Without any priority, rules $r_1$, $r_2$, $r_3$ and $r_4$ are deleted, and the only derived literals are the facts $b, i, j$. If the priorities $r_2 > r_1$ and $r_4 > r_3$ are added to the program, then rules $r_1$ and $r_3$ are deleted and the derived literals are ${\sim}c, b, i, j$, that is the argument for ${\sim}c$ does not reinstate the argument for $a$.

A more formal comparison between our formalism and the approaches cited above is issue of future research.

## 9 Conclusions and Future Work

Defeasible Logic Programming combines Logic Programming and Defeasible Argumentation, and provides the possibility of representing information in the form of defeasible and strict rules in a declarative manner. A query $q$ will be *warranted*, if the argument $\mathcal{A}$ that supports $q$ is found undefeated by the dialectical analysis. During the dialectical analysis certain constrains are imposed for averting different kinds of fallacious argumentation. Thus, DeLP can manage defeasible reasoning and perform contradictory programs.

The defeasible argumentation basis of DeLP allows to build applications that deal with incomplete and contradictory information in dynamic domains, where information may change. Thus, DeLP can be used for representing agent's knowledge and for providing an inference engine. New applications of DeLP are in progress. We expect feedback from them to pursue future extensions.

In (García & Simari, 1999) a model for parallel defeasible logic programming is proposed. Besides existing forms of parallel logic programming, new sources of implicitly exploitable parallelism are considered: building arguments in parallel, searching for defeaters in parallel, and building a dialectical tree in parallel. An implementation of parallel DeLP is in preparation. An extension of DeLP with presumptions is also in study.

The reader may have noticed that the dialectical tree associated with the warrant procedure could become quite large for non trivial situations. Much of the effort expended in the implementation was put on the task of performing an efficient search (García, 2000; Simari *et al.*, 1994a), However, more work should be done.

## Acknowledgments

The authors are grateful to Jürgen Dix, Micheal Gelfond, Ron Loui, Francesca Toni, Hassan Aït-Kaci, Grigoris Antoniou, Simon Parsons, John Pollock, Veronica Dahl and Paul Tarau for many helpful comments and suggestions. We wish to thank especially Carlos Ivan Chesñevar for many helpful discussions and the three anonymous referees for their useful suggestions. This work was partially supported by Secretaría de Ciencia y Técnica Universidad Nacional del Sur.

## References

Alferes, José J., & Pereira, Luis Moniz. (1994). Contradiction: When avoidance equals removal, part I. *Lecture notes in computer science*, **798**, 11–23.

Alferes, José J., Pereira, Luis Moniz, & Przymusinski, Teodor C. (1996). Strong and explicit negation in nonmonotonic reasoning and logic programming. *Lecture notes in computer science*, **1126**, 143–163.

Antoniou, Grigoris, Billington, David, & Maher, Michel J. (1998). Normal forms for defeasible logic. *Pages 160–174 of: Proceedings of international joint conference and symposium on logic programming*. MIT Press.

Antoniou, Grigoris, Maher, Michael J., & Billington, David. (2000a). Defeasible logic versus logic programming without negation as failure. *Journal of logic programming*, **42**, 47–57.

Antoniou, Grigoris, Billington, David, Governatori, Guido, Maher, Michael J., & Rock, Andrew. (2000b). A family of defeasible reasoning logics and its implementation. *Pages 459–463 of: Proceedings of european conference on artificial intelligence (ecai)*.

Antoniou, Grigoris, Billington, David, Governatori, Guido, & Maher, Michael J. (2001). Representation results for defeasible logics. *Acm transactions on computational logic*, **2**(2), 255–287.

Billington, David, De Coster, Koen, & Nute, Donald. (1990). A modular translation from defeasible nets to defeasible logics. *Journal of experimental and theoretical artificial intelligence, 2*, 151–177.

Bondarenko, Andrei, Toni, Francesca, & Kowalski, Robert A. (1993). An assumption-based framework for non-monotonic reasoning. *Proceedings 2nd. international workshop on logic programming and non-monotonic reasoning*, 171–189.

Bondarenko, Andrei, Dung, Phan M., Kowalski, Robert A., & Toni, Francesca. (1997). An abstract, argumentation-theoretic approach to default reasoning. *Artificial intelligence*, **93**, 63–101.

Brewka, Gerhard. (1996). Well-founded semantics for extended logic programs with dynamic preferences. *Journal of artificial intelligence research*, **4**, 19–36.

Brewka, Gerhard. (2001a). Dynamic argument systems: A formal model of argumentation processes based on situation calculus. *Journal of logic and computation*, **11**(2), 257–282.

Brewka, Gerhard. 2001b (Aug.). On the relation between defeasible logic and well-founded semantics. *Proceedings lpnmr 2001*.

Brewka, Gerhard, & Eiter, Thomas. (2000). Prioritizing default logic. *Pages 27–46 of:* Hölldobler, Steffen (ed), *Intellectics and computational logic: Papers in honor of wolfgang bibel*. Dordrecht, Boston, London: Kluwer Academic Publishers.

Chesñevar, Carlos I., Maguitman, Ana G., & Loui, Ronald P. (2000). Logical Models of Argument. *Acm computing surveys*, **32**(4), 337–383.

Chesñevar, Carlos I., Dix, Jürgen, Stolzenburg, Frieder, & Simari, Guillermo R. (2002). Relating defeasible and normal logic programming through transformation properties. *Theoretical computer science*. accepted for publication.

Covington, Michael A., Nute, Donald, & Vellino, Andre. (1997). *Prolog programming in depth*. Prentice-Hall.

Dahl, Veronica. (1999). Logic programming and languages. *Wiley encyclopedia of electrical and electronics engineering*, **11**, 576–580.

Dahl, Veronica, Tarau, Paul, & Li, Renwei. (1997). Assumption grammars for natural language processing. *Fourteenth international conference on logic programming*, 256–270.

Dimopoulos, Yannis, & Kakas, Antonis. (1995). Logic programming without negation as failure. *Pages 369–384 of: Proceedings of 5th. international symposium on logic programming*. Cambridge, MA: MIT Press.

Dix, Jürgen. (1994). Semantics of logic programs: their intuitions and formal properties. *Pages 227–313 of:* Fuhrmann, André, & Rott, Hans (eds), *Logic, action and information*. Berlín–New York: de Gruyter.

Dix, Jürgen, & Stolzenburg, Frieder. (1998). A framework to incorporate non-monotonic reasoning into constraint logic programming. *Journal of logic programming*, **37**, 1–31.

Dung, Phan M. (1991). Negation as hypothesis: An abductive foundation for logic programs. *Proceedings of the 8th. international conference on logic programming*. Paris, France: MIT Press.

Dung, Phan M. (1993a). An argumentation semantics for logic programming with ex-plicit negation. *Pages 616–630 of: Proceedings 10th. intenational conference on logic programming.* MIT Press.

Dung, Phan M. (1993b). On the Acceptability of Arguments and its Fundamental Role in Nomonotonic Reasoning and Logic Programming. *Proceedings of the 13th. international joint conference in artificial intelligence (ijcai), chambéry, francia.*

Dung, Phan M. (1995). On the acceptability of arguments and its fundamental role in nonmonotonic reasoning and logic programming and $n$-person games. *Artificial intelligence*, **77**, 321–357.

Eshghi, Kave, & Kowalski, Robert A. (1989). Abduction compared with negation as failure. *Proceedings of the 6th. international conference on logic programming.* Lisbon, Portugal: MIT Press.

García, Alejandro J. 1997 (July). *Defeasible logic programming: Definition and implementation.* M.Phil. thesis, Computer Science Department, Universidad Nacional del Sur, Bahía Blanca, Argentina.

García, Alejandro J. 2000 (Dec.). *Defeasible logic programming: Definition, operational semantics and parallelism.* Ph.D. thesis, Computer Science Department, Universidad Nacional del Sur, Bahía Blanca, Argentina.

García, Alejandro J., & Simari, Guillermo R. (1999). Parallel defeasible argumentation. *Journal of computer science and technology special issue: Artificial intelligence and evolutive computation. http://journal.info.unlp.edu.ar/*, **1**(2), 45–57.

García, Alejandro J., Simari, Guillermo R., & Chesñevar, Carlos I. 1998 (Aug.). An argumentative framework for reasoning with inconsistent and incomplete information. *Workshop on practical reasoning and rationality.* 13th biennial European Conference on Artificial Intelligence (ECAI-98).

García, Alejandro J., Gollapally, Devender, Tarau, Paul, & Simari, Guillermo R. 2000 (Aug.). Deliberative stock market agents using jinni and defeasible logic programming. *Proceedings of esaw'00 engineering societies in the agents' world, workshop of ecai 2000.*

Gelfond, Michael. (1994). Logic programming and reasoning with incomplete information. *Annals of mathematics and artificial intelligence*, **12**, 89–116.

Gelfond, Michael, & Lifschitz, Vladimir. (1990). Logic programs with classical negation. *Pages 579–597 of:* Warren, D., & Szeredi, P. (eds), *7th international conference on logic programming.* MIT Press.

Gelfond, Michel, & Son, Tran C. (1997). Reasoning with prioritized defaults. *Pages 164–223 of: Lecture notes in artificial intelligence 1471, selected papers from the workshop on logic programming and knowledge representation.*

Inoue, Kazuko. (1991). Extended logic programming with default assumptions. *Proceedings of 8th. international conference on logic programming.*

Kakas, Antonis C., Kowalski, Robert A., & Toni, Francesca. (1993). Abductive logic programming. *Journal of logic and computation*, **2**, 719–770.

Kakas, Antonis C., Mancarella, Paolo, & Dung, Phan M. (1994). The acceptability semantics for logic programs. *Pages 504–519 of: Proceedings of the 11th. international conference on logic programming.* Santa Margherita, Italy: MIT Press.

Kowalski, Robert A., & Toni, Francesca. (1996). Abstract argumentation. *Artificial intelligence and law*, **4**(3-4), 275–296.

Li, Renwei, Pereira, Luis Moniz, & Dahl, Veronica. (1998). Refining action theories with abductive logic programming. *Selected extended papers from the lpkr'97: Ilps'97 workshop on logic programming and knowledge representation*, 123–138.

Lifschitz, Vladimir. (1996). Foundations of logic programs. *Pages 69–128 of:* Brewka, G. (ed), *Principles of knowledge representation.* CSLI Pub.

Loui, Ronald P. (1997a). Alchourrón and Von Wright on Conflict among Norms. *Pages 345–353 of:* Nute, Donald (ed), *Defeasible deontic logic*, vol. 263. Synthese Library.

Loui, Ronald P. 1997b (July). et al. Progress on Room 5: A Testbed for Public Interactive Semi-Formal Legal Argumentation. *Proceedings of the 6th. international conference on artifcial intelligence and law.*

Makinson, David, & Schlechta, Karl. (1991). Floating conclusions and zombie paths: two deep difficulties in the directly skeptical approach to defeasible inference nets. *Artificial intelligence*, **48**, 199–209.

Nute, Donald. (1988). Defeasible reasoning: a philosophical analysis in PROLOG. *Pages 251–288 of:* Fetzer, J. H. (ed), *Aspects of artificial intelligence*. Kluwer Academic Pub.

Nute, Donald. (1992). Basic defeasible logic. Fariñas del Cerro, Luis (ed), *Intensional logics for programming*. Oxford: Claredon Press.

Nute, Donald. (1994). Defeasible logic. *Pages 355–395 of:* Gabbay, D.M., Hogger, C.J., & J.A.Robinson (eds), *Handbook of logic in artificial intelligence and logic programming, vol 3*. Oxford University Press.

Pereira, Luis Moniz, & Alferes, José J. (1994). Contradiction: When avoidance equals removal, part II. *Lecture notes in computer science*, **798**, 268–281.

Pollock, John. (1987). Defeasible Reasoning. *Cognitive science*, **11**, 481–518.

Pollock, John. (1995). *Cognitive carpentry: A blueprint for how to build a person*. MIT Press.

Pollock, John. (1996). Oscar - A general purpose defeasible reasoner. *Journal of applied non-classical logics*, **6**, 89–113.

Poole, David L. (1985). On the Comparison of Theories: Preferring the Most Specific Explanation. *Pages 144–147 of: Proc. 9th IJCAI*. IJCAI.

Prakken, Henry. (1997). *Logical tools for modelling legal argument. a study of defeasible reasoning in law*. Kluwer Law and Philosophy Library.

Prakken, Henry, & Sartor, Giovanni. (1997). Argument-based logic programming with defeasible priorities. *J. of applied non-classical logics*, **7**(25-75).

Prakken, Henry, & Vreeswijk, Gerard. (2000). Logical systems for defeasible argumentation. D.Gabbay (ed), *Handbook of philosophical logic, 2nd ed.* Kluwer Academic Pub.

Simari, Guillermo R., & Loui, Ronald P. (1992). A Mathematical Treatment of Defeasible Reasoning and its Implementation. *Artificial intelligence*, **53**, 125–157.

Simari, Guillermo R., Chesñevar, Carlos I., & García, Alejandro J. 1994a (Oct.). Focusing inference in defeasible argumentation. *Iv iberoamerican congress on artificial intelligence (iberamia'94)*. Venezuela.

Simari, Guillermo R., Chesñevar, Carlos I., & García, Alejandro J. 1994b (Nov.). The role of dialectics in defeasible argumentation. *XIV international conference of the chilenean computer science society*.

Toni, Francesca, & Kakas, Antonis C. (1995). Computing the acceptability semantics. *Pages 401–415 of: Proceedings of the 3rd. international workshop on logic programming and non-monotonic reasoning*. Lexington,USA: Springer Verlag.

Verheij, Bart. 1996 (Dec.). *Rules, reasons, arguments: formal studies of argumentation and defeat*. Ph.D. thesis, Maastricht University, Holland.

Vreeswijk, Gerard A. W. (1997). Abstract argumentation systems. *Artificial intelligence*, **90**, 225–279.

Xianchang Wang, Jia-Huai You, Li Y. Yuan. (1997). Logic programming without default negation revisited. *Pages 1169–1174 of: Proceedings of ieee international conference on intelligent processing systems*. IEEE.